\documentclass{article}
\usepackage[utf8]{inputenc}
\usepackage{authblk}
\usepackage{setspace}
\usepackage[margin=1.25in]{geometry}
\usepackage{graphicx}
\graphicspath{ {./figures/} }
\usepackage{subcaption}
\usepackage{amsmath}
\usepackage{lineno}
\usepackage{booktabs}
\usepackage{array}
\usepackage{tabularx}
\usepackage{rotating}
\usepackage{xcolor}
\usepackage{tikz}
\usetikzlibrary{arrows.meta,positioning,fit,backgrounds}
\usepackage[hidelinks]{hyperref}
\newif\ifdraft
\draftfalse
\ifdraft
  
\else
  
\fi
\newcommand{\figplaceholder}[2]{%
  \fbox{\parbox[c][#1][c]{0.95\linewidth}{\centering\small\textcolor{red!70!black}{\textbf{[FIGURE TBD]} #2}}}}
\newcommand{\code}[1]{\texttt{\small #1}}

\usepackage[style=nejm,
citestyle=numeric-comp,
sorting=none,
backend=bibtex]{biblatex}
\title{From Video to Digital Twin: An Exploratory Study of World Models and Gaussian Splatting on In-the-Wild Video}

\author[1*]{Somesh Mohapatra}

\affil[1]{Independent Researcher.}
\affil[*]{Address correspondence to: pikulsomesh@gmail.com}

\date{}

\begin{document}

\maketitle

%%%%%% Abstract %%%%%%
\begin{abstract}
Generative world models and 3D Gaussian Splatting promise to turn ordinary
video into explorable 3D scenes, but little is reported on how they behave on
uncurated, in-the-wild footage. We study what these models recover when the
only input is a publicly available news walkthrough of a food bank warehouse,
with presenters walking through every shot and no drawings, scans or site
visit, and we use the result as the geometry of a digital twin. A commercial
world model given 20 frames recovered camera poses and metric depth for three
areas, which we fused into about 747,000 Gaussian splats for 200 credits (20
US cents). Splat models optimized locally on a CPU, at no credit cost, were
photorealistic near the filmed path but degraded quickly away from it. On
held-out frames, scored in pixels not covered by people, they reached 21.2 and
23.1 dB PSNR from 64 and 43 training frames, and lost 3.8 and 5.4 dB when
given only the 8 frames a world model accepts. Removing the trainer's background fill and person-hull
carving raised held-out PSNR, showing that cleaning the space behind moving
people costs fidelity in observed pixels while its gain lies where no ground
truth exists. Asking the world model to generate views of areas the video
never showed failed on the provider's side twice, so extrapolation could not
be evaluated and unseen space was filled by a scripted model instead. We also
describe the corrections needed to turn world-model output into a valid,
metric Universal Scene Description (OpenUSD) asset with its native splat
schema, which serves as the only
interface to a data layer, a browser viewer, Blender and NVIDIA Omniverse. The
findings suggest that today's models recover the observed parts of a real
scene cheaply and at metric scale, while few-view fidelity and extrapolating
unobserved space remain open. The work is an exploration of these methods,
not a deployed product.
\end{abstract}

%%%%%% Main Text %%%%%%

\section{Introduction}
A digital twin is a virtual representation of a physical asset that stays
synchronized with it through data \cite{grieves2017digital,kritzinger2018digital}.
In industry, twins are used to plan layouts, monitor equipment and rehearse
changes before they are made on the floor \cite{tao2019digital,fuller2020digital}.
The geometry of a facility twin is usually the expensive part. It comes from
terrestrial laser scanning, photogrammetry campaigns or hand-built
computer-aided design (CAD) models, and specialists then import it into a
simulation tool. That cost is tolerable for an automotive plant. It is
prohibitive for a regional food bank, which runs warehouse-scale cold-chain
logistics on a non-profit budget.

Two developments change that cost structure. Neural scene representations,
and 3D Gaussian Splatting in particular \cite{kerbl20233d}, produce
photorealistic scenes that render in real time from ordinary images. Classical
reconstruction recovers camera poses with structure-from-motion
\cite{schoenberger2016sfm}; neural radiance fields \cite{mildenhall2020nerf}
improved view synthesis but render slowly; learned multi-view geometry such as
DUSt3R \cite{wang2024dust3r} reduces the number of views needed. All of these
reconstruct only what was observed. Generative world models
\cite{ha2018world,brooks2024video} go further: from a few views they produce a
coherent, explorable 3D world and fill in regions the input never covered.
World Labs' Marble \cite{worldlabs_marble} exposes this capability through an
API that returns Gaussian-splat worlds. For a twin, generation has a double
edge. It makes a sparse walkthrough usable, but it also invents geometry.

Three lines of recent work bear directly on footage like ours. First,
feed-forward geometry models predict poses and dense geometry from a handful of
unposed images: DUSt3R \cite{wang2024dust3r}, MASt3R \cite{leroy2024mast3r} and
VGGT \cite{wang2025vggt}. Sparse-view splatting methods such as FSGS
\cite{zhu2023fsgs} and InstantSplat \cite{fan2024instantsplat} build on such
priors to train Gaussian splats from a few views. Second, in-the-wild
reconstruction handles transient content, from appearance embeddings and
transient masks for radiance fields (NeRF in the Wild \cite{martinbrualla2020nerfw})
to distractor-robust splatting (WildGaussians \cite{kulhanek2024wildgaussians},
SpotLessSplats \cite{sabour2024spotless}); the presenters who walk with the
camera in our video are exactly such transients. Third, generative models
synthesize unobserved views: multi-view diffusion (CAT3D \cite{gao2024cat3d}),
video diffusion conditioned on a point cloud (ViewCrafter \cite{yu2024viewcrafter})
and interactive world models trained on video (Genie \cite{bruce2024genie}).
Faster and anti-aliased radiance fields \cite{muller2022instantngp,barron2022mipnerf360}
set the evaluation conventions we follow: novel-view metrics on held-out frames.
We do not benchmark these methods; we use a commercial world model and a
standard 3DGS objective as representatives and measure what they recover from
one uncurated video.

In parallel, Universal Scene Description (OpenUSD), originally from Pixar and
now developed under the Alliance for OpenUSD, has become an open,
vendor-neutral scene description with layered composition and typed schemas
\cite{openusd,aousd}. Its 26.03 release added the
\code{ParticleField3DGaussianSplat} schema family
\cite{aousd2603,usd_gsplat_schema}, so a splat reconstruction is no longer a
viewer-specific file. It can sit in the same stage as sensors, equipment and
annotations. NVIDIA Omniverse \cite{omniverse} is a set of libraries that bring
OpenUSD data interoperability, RTX rendering and physics simulation into
applications. Its licensing is mixed: some
components are open source, while the Kit SDK is distributed under NVIDIA's
own license terms: the Kit App Template states that its software and materials
are governed by the NVIDIA Software License Agreement and the Product-Specific
Terms for NVIDIA Omniverse \cite{kit_app_template}, not an open-source
license.

Surveys of digital twins identify geometry acquisition, data integration and
interoperability as the recurring barriers \cite{tao2019digital,fuller2020digital}.
This work addresses the first and third barriers directly and keeps the second
deliberately thin. This paper is an exploratory study of the learning
methods behind the first barrier. We ask three questions: what a world model
and locally optimized Gaussian splats recover from uncurated, publicly
available video with no CAD drawings, laser scans or site visit; whether the
generative step can fill in space the camera never saw; and what it takes to
carry the result into an open scene format that other tools can use. To
answer them we compose existing models into a three-step pipeline
(Figure~\ref{fig:overview}) and apply it to one realistic input, a news
walkthrough of a food bank. The steps are:
\begin{enumerate}
  \item \textbf{Video to digital twin with world models.} Blur-aware frame
        extraction feeds a world model that returns a Gaussian-splat world,
        which is converted to an OpenUSD stage in metric, Y-up coordinates.
        Where the camera saw too little, a facility model scripted in Blender
        fills in the building, published in the same format.
  \item \textbf{Connecting existing data systems on Amazon Web Services (AWS).}
        A serverless API exposes operational data in the twin's coordinate
        frame through a small, versioned JSON contract. The reference
        implementation simulates cold-chain telemetry; the contract is
        designed so that a real sensor feed or warehouse management system
        can replace the simulator without changing any renderer.
  \item \textbf{NVIDIA Omniverse as the visualization layer.} Omniverse
        renders the USD twin with RTX on a rented GPU; binding live data to
        scene objects is the planned next step. A
        browser viewer and Blender read the same USD and the same JSON, so
        the renderer is a deployment choice rather than an architectural one.
\end{enumerate}

We do not propose a new model; the contribution is an empirical account of
how current models behave on in-the-wild input. It has three parts: held-out
novel-view scores for Gaussian splats trained on the footage, with the number
of input frames and the handling of people who walk with the camera varied
(Table~\ref{tab:heldout}); a qualitative comparison with a world-model capture
from eight frames, including where each fails; and a report that the
world model's generative step could not be evaluated because it failed on
the provider's side. The engineering needed to use the result (coordinate and
packaging fixes that make world-model output a valid metric OpenUSD asset, a
data backend, viewers and an operational technology security design) is
summarized in Results and detailed in Appendix~\ref{app:system}. Food banks are logistics operations: when
the Central Texas Food Bank moved into its 135,000-square-foot Austin
facility in 2016, the new building held seven times the refrigeration and
freezer capacity of the old one \cite{ctfb2016facility}, and its newer Waco
hub combines a warehouse and distribution hub, a kitchen, a food market and a
training center in 64,000 square feet \cite{kxxv2026waco}. Cold storage at that
scale is exactly what a twin with live temperatures is meant to watch.

The case study is a representative Texas food bank, not a model of any one
facility. It is inspired by the Central Texas Food Bank: three of its areas
are captured from a public walkthrough of one of that food bank's buildings,
and its operating volumes are anchored to published figures for another. The
rest of the building is designed for this paper, and all live data is
simulated. Table~\ref{tab:provenance} states where every part of the twin
comes from, and the figures show captured, designed and simulated content so
that they can be told apart. The aim is to show what a food bank's twin looks
like and what it takes to build one, not to describe a particular site.

\section{Results}

\subsection{Three steps joined by one open contract}
\begin{figure}[t]
\centering
\begin{tikzpicture}[
  box/.style={draw, rounded corners=2pt, align=center, font=\small, minimum height=1.3cm, text width=2.85cm, fill=white},
  lay/.style={draw=black!25, rounded corners=4pt, inner xsep=6pt, inner ysep=5pt, fill=black!4},
  lbl/.style={font=\footnotesize\bfseries, anchor=south west, inner sep=1pt},
  arr/.style={-{Stealth[length=2.2mm]}, thick},
  node distance=0.9cm and 0.45cm]
  % Step 1
  \node[box] (vid) {Walkthrough video\\\footnotesize MP4 or phone};
  \node[box, right=of vid] (frm) {Blur-aware frames\\\footnotesize ffmpeg, $\le$40};
  \node[box, right=of frm] (wm) {World model\\\footnotesize Gaussian-splat world};
  \node[box, right=of wm] (usd) {OpenUSD stage\\\footnotesize metric, Y-up};
  \node[lbl] at ([yshift=3pt]vid.north west) (l1) {Step 1: video $\rightarrow$ twin with a world model};
  \draw[arr] (vid) -- (frm); \draw[arr] (frm) -- (wm); \draw[arr] (wm) -- (usd);
  % Step 2 (columns 1-3, leaving column 4 free for the USD arrow)
  \node[box, below=1.25cm of vid] (src) {Existing systems\\\footnotesize IoT, WMS, cold chain};
  \node[box, right=of src] (api) {Serverless API\\\footnotesize Lambda + HTTP API};
  \node[box, right=of api] (json) {Telemetry JSON\\\footnotesize meters, twin frame};
  \node[lbl] at ([yshift=3pt]src.north west) (l2) {Step 2: existing data on AWS};
  \draw[arr] (src) -- (api); \draw[arr] (api) -- (json);
  % contract bar
  \coordinate (barl) at ([yshift=-1.1cm]src.south -| vid.west);
  \coordinate (barr) at ([yshift=-1.1cm]src.south -| usd.east);
  \node[draw, thick, rounded corners=2pt, fill=blue!6, fit=(barl)(barr), inner xsep=0pt, inner ysep=0.35cm,
        label={[font=\small]center:Open contract: OpenUSD splat stage, SPZ and metadata, plus the telemetry JSON}] (bar) {};
  \draw[arr] (usd.south) -- (usd.south |- bar.north);
  \draw[arr] (json.south) -- (json.south |- bar.north);
  % Step 3
  \node[box, below=0.9cm of bar.south, anchor=north] (ov) {NVIDIA Omniverse\\\footnotesize RTX + live-data extension};
  \node[box, left=of ov] (web) {Browser\\\footnotesize Spark on SPZ};
  \node[box, right=of ov] (bl) {Blender\\\footnotesize alternative};
  \node[lbl, anchor=north west] at ([yshift=-3pt]web.south west) (l3) {Step 3: visualization};
  \foreach \n in {web,ov,bl} \draw[arr] (bar.south -| \n.north) -- (\n.north);
  \begin{scope}[on background layer]
    \node[lay, fit=(l1)(vid)(usd)] {};
    \node[lay, fit=(l2)(src)(json)] {};
    \node[lay, fit=(l3)(web)(bl)] {};
  \end{scope}
\end{tikzpicture}
\caption{The pipeline. The open contract is the only
interface between steps: the OpenUSD stage (with the SPZ it was converted from
and a metadata file) and the telemetry JSON. Any box can be replaced without
changing the others.}
\label{fig:overview}
\end{figure}
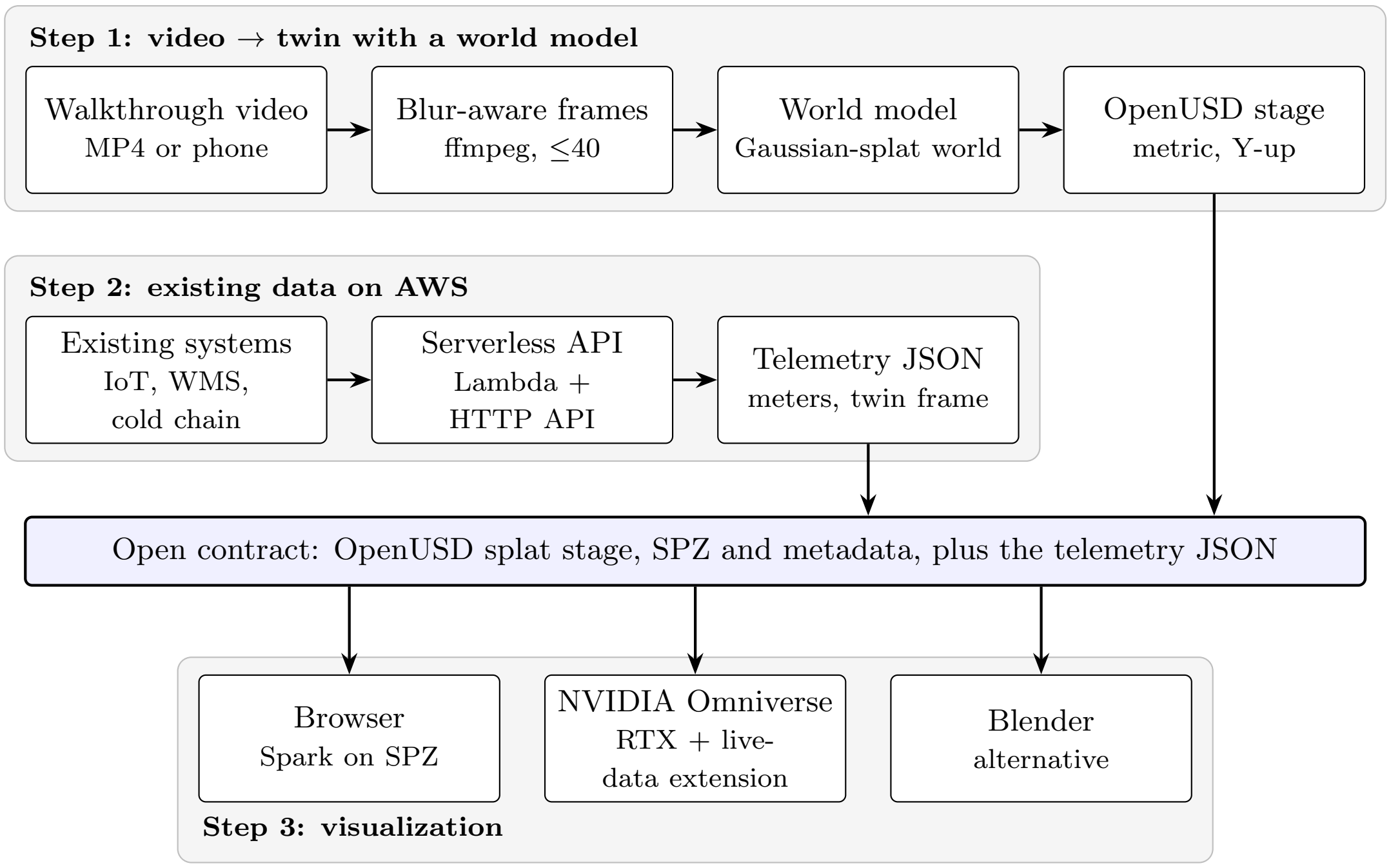

The steps share data, not code. Two artifacts cross step boundaries
(Table~\ref{tab:contracts}). The reconstruction step publishes a USD package,
the SPZ splat file it was converted from (for lightweight viewers) and a
metadata file recording scale, ground offset and provenance. The data step
publishes a JSON document whose positions are in meters in the same frame as
the USD stage. A renderer needs nothing else, which is what lets Omniverse, a
browser and Blender coexist (Figure~\ref{fig:overview}).

\begin{table}[t]
\caption{Contracts between the pipeline's steps. Changes to the telemetry contract are additive only.}
\centering\small
\begin{tabularx}{\linewidth}{@{}l>{\raggedright\arraybackslash}p{3.6cm}>{\raggedright\arraybackslash}X@{}}
\toprule
Boundary & Artifact & Rules \\
\midrule
1 to 3 & \code{factory.usdz}, \code{factory.spz}, \code{factory.meta.json} &
  Metric (\code{metersPerUnit}\,=\,1), Y-up, ground at $y=0$; alignment and provenance in the metadata file. \\
2 to 3 & \code{GET /telemetry} (JSON) &
  \code{zones[]} (temperature, setpoint, humidity, door, status, position) and
  \code{forklifts[]} (position, heading, speed, battery); positions in meters in the twin frame; \code{simulated} flag. \\
\bottomrule
\end{tabularx}
\label{tab:contracts}
\end{table}

\subsection{A walkthrough news video yields usable input for three zones}
The input was a public news walkthrough of a new Central Texas Food Bank
facility, which by its broadcaster and on-screen campaign appears to be the
Waco hub \cite{kxxv2026waco} (18\,min\,32\,s, 1920\,$\times$\,1080 at 59.94 frames per second)
\cite{kxxv2026ctfb}. Two people appear in almost every shot and a lower-third
graphic is burned into each frame, which makes it a realistic rather than a
curated input. A multi-image world needs overlapping views of one space, so we
targeted a single zone. Surveying the video at one frame every 10\,s gave the
candidate zones in Table~\ref{tab:zones}. The pallet-racking aisle was
captured first, because the camera circles one bay and then looks down the
aisle, and because it is the zone that matters most for inventory locations,
forklift paths and telemetry. The volunteer sorting room and the client market
were captured in a second run, from narrower windows in which the camera moves
enough to give overlapping views. The other operating areas of the
representative food bank (receiving, freezer, cooler and order staging) were
not captured: they are designed footprints in the operations layer, drawn by
a modeled facility, not generated or reconstructed geometry
(Table~\ref{tab:provenance}). The community kitchen was captured later by
local training only.

\begin{table}[t]
\caption{Provenance of each part of the representative food bank twin.}
\centering\small
\begin{tabularx}{\linewidth}{@{}>{\raggedright\arraybackslash}p{0.3\linewidth}>{\raggedright\arraybackslash}p{0.17\linewidth}>{\raggedright\arraybackslash}X@{}}
\toprule
Part of the twin & Kind & Source \\
\midrule
Racking aisle, sorting room, client market (geometry) & Captured, partial & 20 frames of a public walkthrough of a Central Texas Food Bank building; world-model depth fused locally. The same three are also trained locally from 49 to 72 frames each (Table~\ref{tab:photoreal}). \\
Community kitchen interior (geometry) & Captured, partial & Trained locally from 65 frames of the same video; no world-model capture. \\
Absolute scale & Estimated & The world model's metric depth; not checked against a survey. \\
Building shell, receiving, freezer, cooler, order staging, kitchen annex footprint & Designed & Footprints chosen for this paper. \\
Modeled facility (racking, cold rooms, sorting lines, market, kitchen equipment, forklifts) & Designed & Built in Blender by one script from the designed footprints, with fit-out inspired by the video; a compact 26\,$\times$\,20\,m building plus an 11.5\,$\times$\,11\,m annex, much smaller than the facility whose volumes drive the operations model. \\
Annual volume, agencies, floor area, pallet positions, cold storage, docks, kitchen equipment, meals per day & Published & Figures for the Central Texas Food Bank's Austin facility (Table~\ref{tab:published}). \\
Source and channel shares, product mix, capacities, recipes, step times & Estimated & Set for this paper and flagged as estimates in the master data. \\
Temperatures, door events, forklifts, daily flows and inventory & Simulated & Computed from the clock; every payload carries \code{simulated: true}. \\
\bottomrule
\end{tabularx}
\label{tab:provenance}
\end{table}

\begin{table}[t]
\caption{Candidate zones in the source walkthrough and the assessment that led to the three captured zones.}
\centering\small
\begin{tabularx}{\linewidth}{@{}llX@{}}
\toprule
Time (min:s) & Zone & Assessment \\
\midrule
0:45--1:35 & Volunteer sorting room & Captured from 0:48--1:06, where the camera moves; 6 frames. \\
2:10--2:50 & Freezer and cooler doors & Mostly blank walls; fogged at 2:40. \\
5:05--6:00 & Loading dock & People fill the frame. \\
6:10--7:45 & Pallet-racking aisle & Captured. Racking, wrapped pallets and a pallet jack; the camera circles a bay and looks down the aisle; one view has no people; 8 frames. \\
9:20--10:00 & Client grocery market & Captured from 9:16--9:50; people block the center and were masked; 6 frames. \\
13:50--16:40 & Commercial kitchen & Reflective stainless steel, hard for splats. \\
\bottomrule
\end{tabularx}
\label{tab:zones}
\end{table}

\begin{figure}[t]
\centering
\IfFileExists{figures/contact_sheet.jpg}{%
  \includegraphics[width=\linewidth]{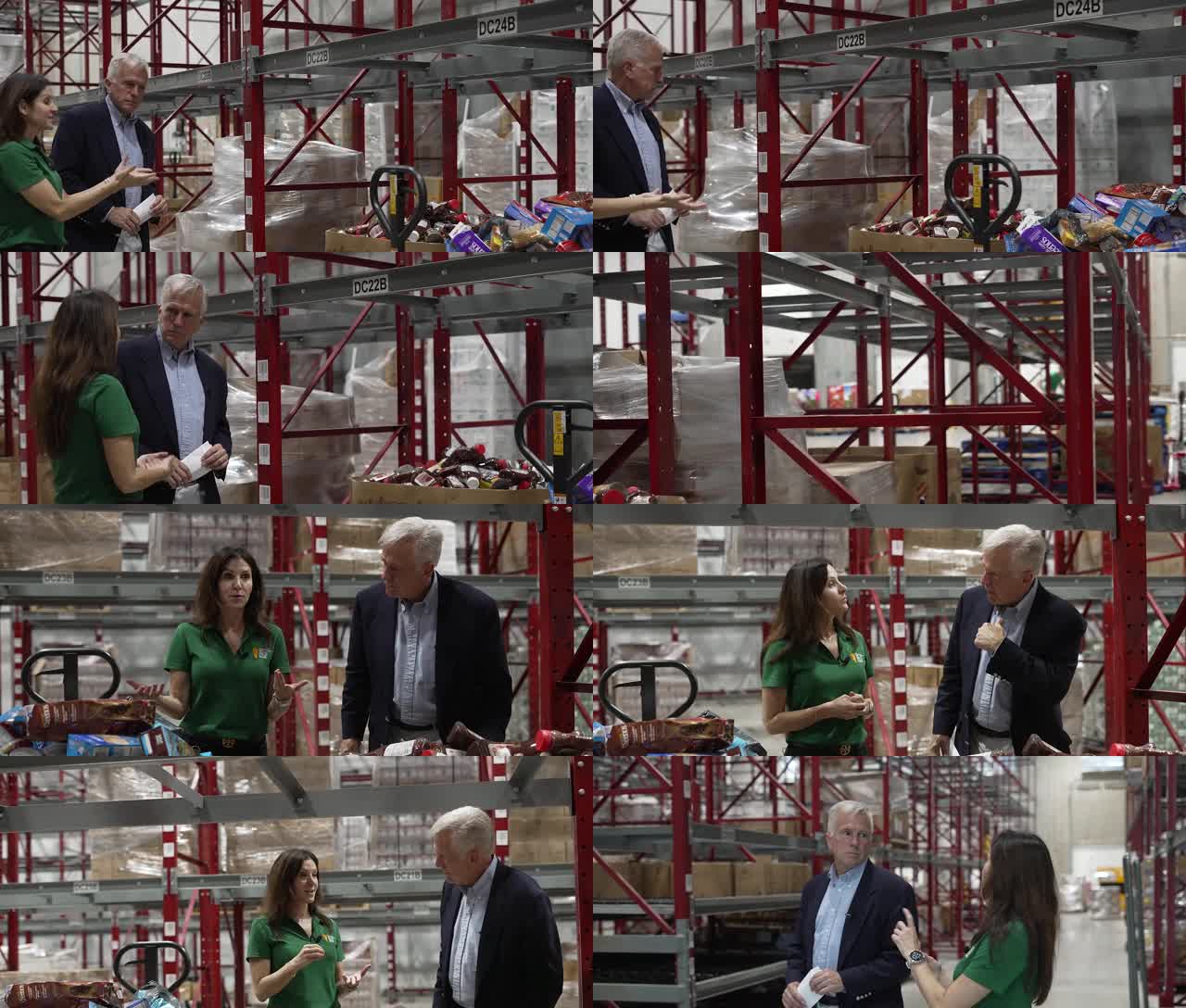}}{%
  \figplaceholder{4cm}{Contact sheet of the eight input frames.}}
\caption{The eight input frames from the racking aisle, 1920\,$\times$\,816
pixels after cropping the broadcast graphic. The sorting room and market
inputs are in the repository.}
\label{fig:inputs}
\end{figure}

Automatic frame selection failed on this footage. Ranking frames by variance
of the Laplacian, a standard sharpness score, rewards in-focus faces, and on
this video it selected six near-identical shots of the presenters. The eight
inputs were therefore chosen by hand for viewpoint spread and absence of
people, spanning 6:13 to 7:33 (Figure~\ref{fig:inputs}): three views of one bay
from different angles, one view with no people, two from the opposite side of
the aisle, a wide shot under the beam and one looking down the main aisle
toward the dock. The sorting room and market inputs, six frames each, were
picked by hand in the same way from 40 extracted frames per zone. In every
input, the two presenters were covered by hand-drawn boxes that exclude those
pixels from reconstruction.

\subsection{The world model recovers metric geometry, but did not yet extrapolate}
The world model posed every input frame and returned metric depth with a
per-pixel confidence. The depth is plausible at real scale: in the racking
aisle, the first beam lands about 0.4\,m above the camera, consistent with a
beam about 1.9\,m above the floor for a handheld camera. Fusing the posed
frames gave 270{,}423 splats for the racking aisle, about 282{,}000 for the
sorting room (keeping every third pixel) and about 195{,}000 for the client
market, about 747{,}000 in total (Table~\ref{tab:results}). The floor was found from a dense horizontal
slab of points in the sorting room and from the world model's own ground plane
in the market; in the racking aisle, where the floor is barely visible, it was
placed 1.5\,m below the source cameras.

\begin{figure}[t]
\centering
\begin{subfigure}{0.49\textwidth}
  \includegraphics[width=\linewidth]{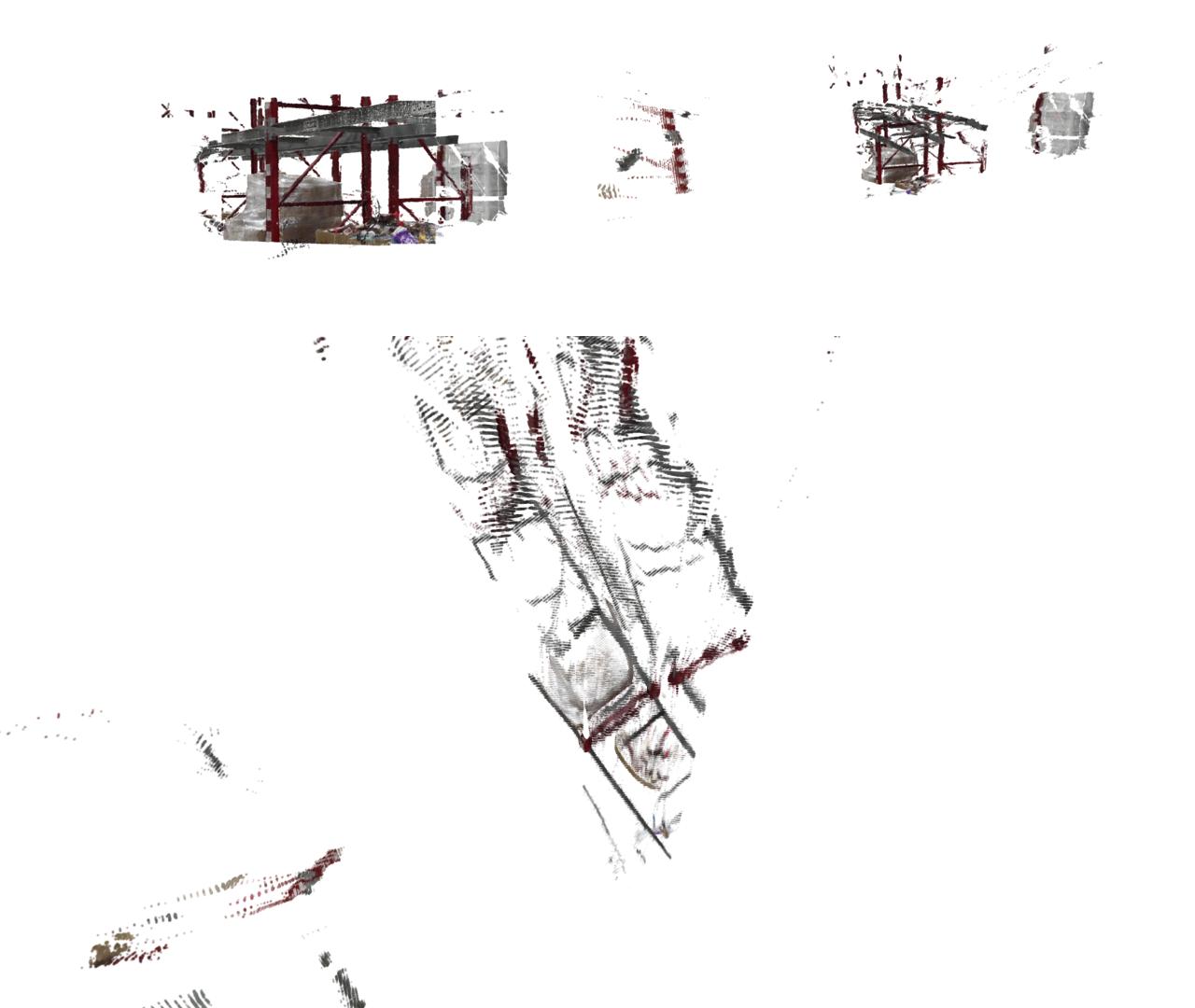}
  \caption{\label{fig:resultsA}}
\end{subfigure}\hfill
\begin{subfigure}{0.49\textwidth}
  \includegraphics[width=\linewidth]{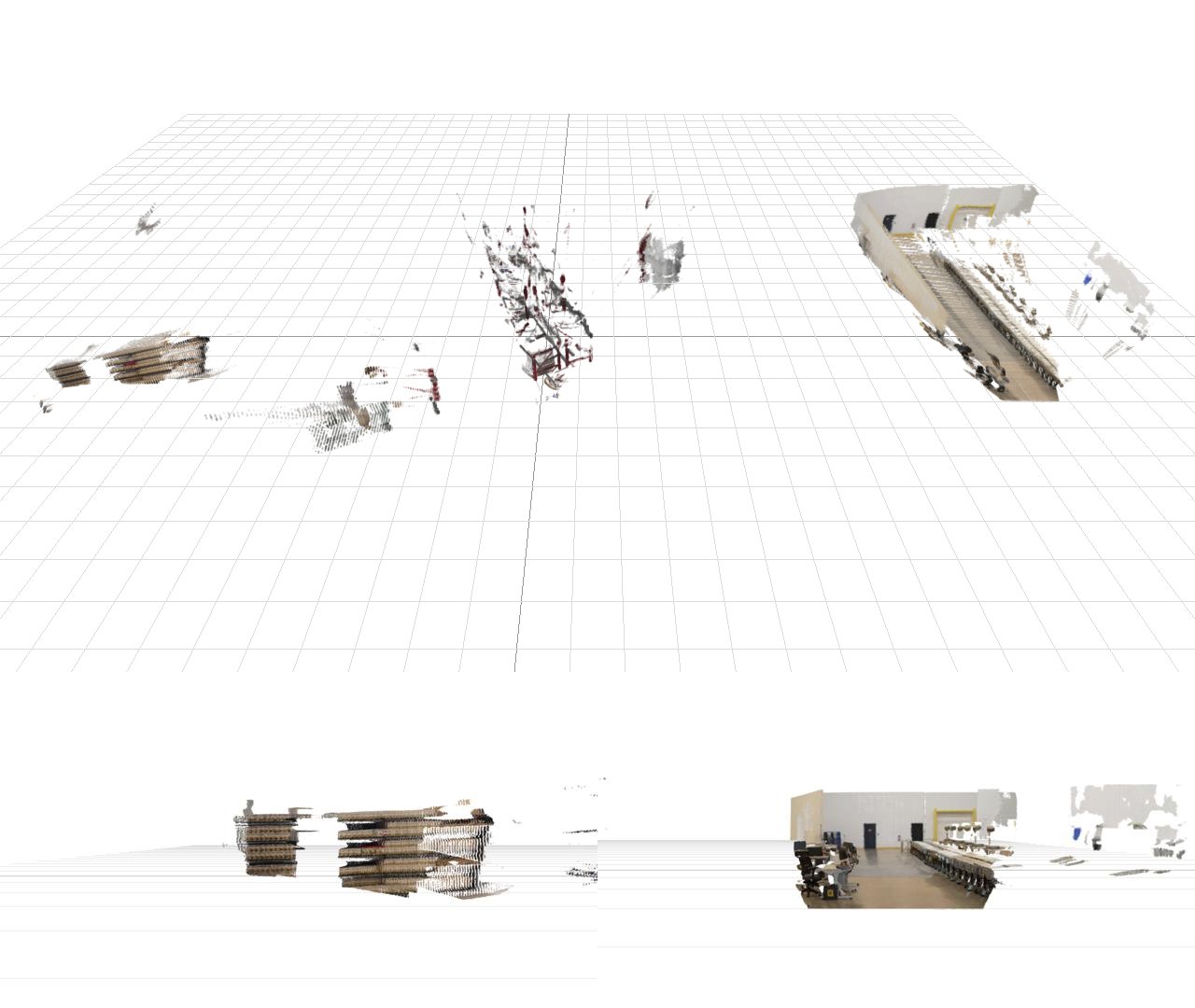}
  \caption{\label{fig:resultsB}}
\end{subfigure}
\caption{The fused twin rendered in Spark 0.1.10, the viewer version at the time of capture (the current viewer uses Spark 2.2). (\subref{fig:resultsA}) The
racking aisle from two viewpoints: red racking, a wrapped pallet and the
donation bin are recovered where the camera looked, and the rest of the aisle
is empty. (\subref{fig:resultsB}) Top: the three zones placed on one site plan
on a 1\,m grid; bottom: the market and the sorting room from their source
viewpoints. No generated views are included.}
\label{fig:results}
\end{figure}

The generative step did not succeed. Both requests to generate a ring of eight
new views around the racking aisle returned \code{EXECUTION\_FAILED} without
further detail: the first placed every target camera at one center, the
second spread them on a 0.6\,m ring and centered the context principal points.
Retries were paused until it is known whether failed requests are billed. The
twin therefore covers only what the camera saw (Figure~\ref{fig:results}):
each zone is a partial shell seen from its source viewpoints, with gaps where
the presenters were masked and where the camera never pointed. These gaps are
exactly where generated views are meant to go.

\subsection{Locally trained splats make four zones photorealistic at no credit cost}
\begin{figure}[t]
\centering
\includegraphics[width=\linewidth]{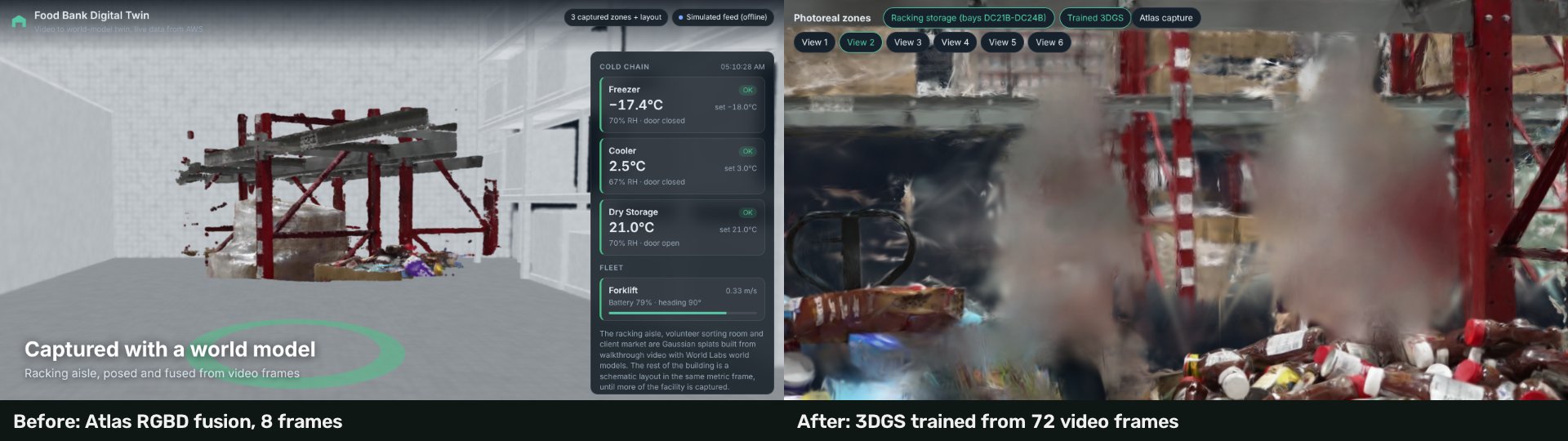}
\caption{The racking aisle in the browser viewer. Left: the world-model capture,
posed and fused from 8 frames. Right: a 3D Gaussian Splatting model trained
locally from 72 frames of the same video, from a source camera, in the first
training run; the grey haze is where the masked presenters stood, which the
background fill and hull carving described above later reduced to a soft blur.}
\label{fig:photoreal}
\end{figure}

To separate what the world model contributes from what Gaussian splatting
contributes, the racking aisle was also reconstructed the conventional way,
without any paid call. A 100\,s segment of the tour (from 6:08) was sampled at
4 frames per second and 960 pixels wide; people were masked automatically,
together with the broadcast graphic; and sharp frames in which the camera had
moved were kept, leaving 72 frames, nine times as many as the world-model
capture used. Camera poses came from structure-from-motion on the CPU
\cite{schoenberger2016sfm,colmap}, and a 3D Gaussian Splatting model
\cite{kerbl20233d} was trained on the CPU with a trainer written for this work.
Near the filmed path, the racking, wrapped pallets, shelf labels and the bottle
bin read photographically, where the world-model capture shows sparse
fragments (Figure~\ref{fig:photoreal}).

\begin{figure}[t]
\centering
\includegraphics[width=\linewidth]{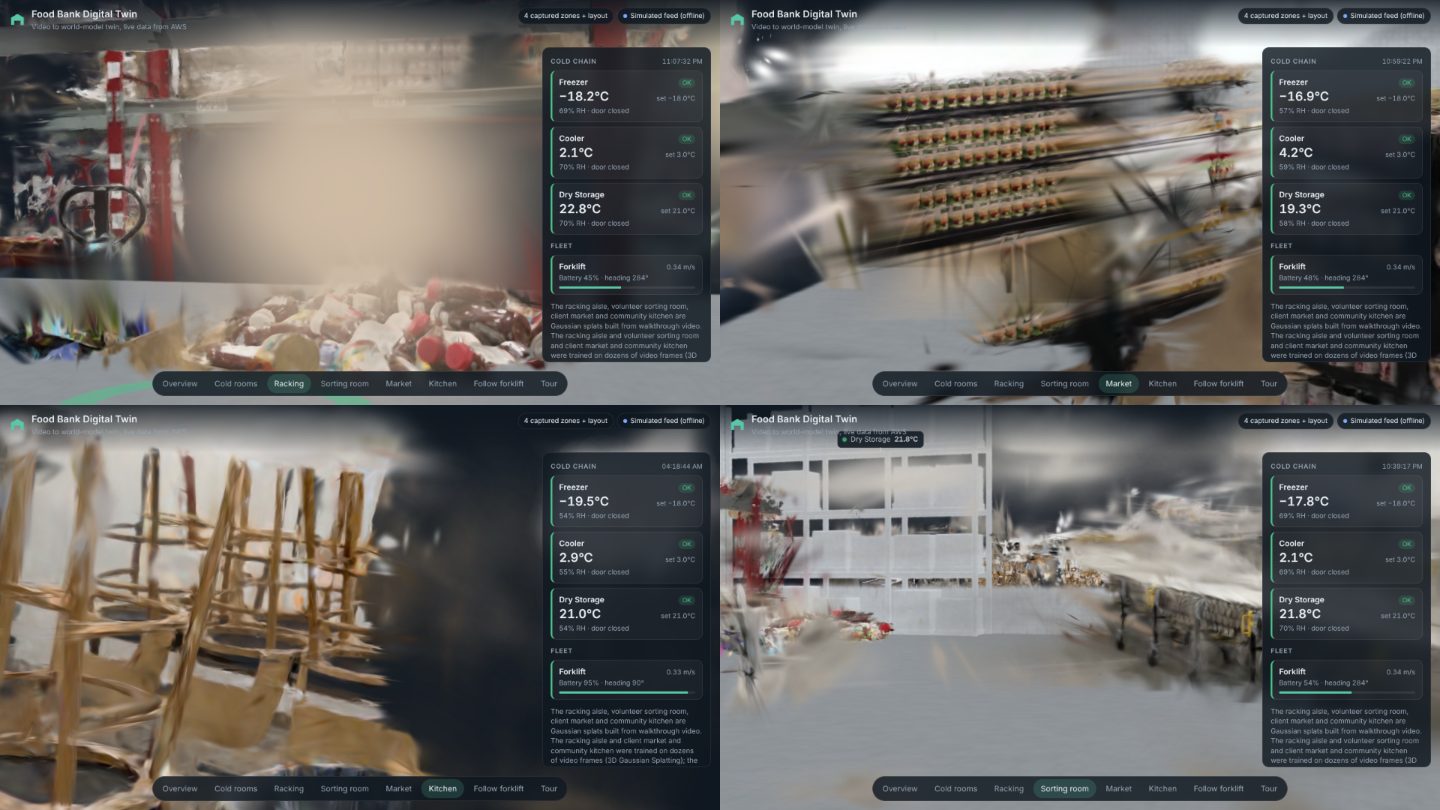}
\caption{The four locally trained zones in the browser viewer, each from one
of its source cameras: racking aisle (top left), client market (top right),
community kitchen (bottom left) and volunteer sorting room (bottom right). The
panels show simulated telemetry. Soft regions are where the presenters stood or
the camera did not look.}
\label{fig:photoreal4}
\end{figure}

The same route was then run on the sorting room, the client market and the
community kitchen, which the world model never captured (Table~\ref{tab:photoreal}).
The four zones hold 105{,}899 to 161{,}425 splats in 1.6 to 2.5\,MB SPZ files
and 6.9 to 10.6\,MB USDZ packages, with the same contract as the world-model
zones, and cost 0 credits. Two changes to the trainer address the
presenters, who walk with the camera, so the space behind them is masked in
nearly every frame: the trainer fills the masked pixels with background warped
in from frames where they had moved, and every 250 steps it removes splats
that sit inside their visual hull. By eye they turn the haze where the
presenters stood into a soft blur; on held-out frames they cost fidelity in
the pixels that were observed (Table~\ref{tab:heldout}). The
sorting room, white walls over a grey floor, needed exhaustive feature
matching before structure from motion would converge. In the browser, all four
zones sit in the representative building, the kitchen in its annex
(Figure~\ref{fig:photoreal4}). Three trade-offs remain. Structure from
motion has no metric scale, so each scene was scaled by assuming the camera
was held 1.6\,m above the floor, where the world model measured depth
directly. The spot right behind the presenters was never filmed and trains to
a soft blur rather than to holes or ghosts. And quality falls off quickly away
from the filmed path, because the walkthrough moves mostly forward.

\begin{table}[t]
\caption{Locally trained zones. Frames are the posed frames recorded in each
zone's export metadata; sizes are the committed files. All cost 0 credits.}
\centering\small
\begin{tabular}{@{}lrrrrr@{}}
\toprule
Zone & Video segment & Frames & Splats & SPZ (MB) & USDZ (MB) \\
\midrule
Racking aisle & 6:08, 100\,s & 72 & 128{,}884 & 2.1 & 8.4 \\
Volunteer sorting room & 0:52, 30\,s & 68 & 111{,}400 & 1.8 & 7.3 \\
Client market & 9:13, 45\,s & 49 & 105{,}899 & 1.6 & 6.9 \\
Community kitchen & 14:23, 34\,s & 65 & 161{,}425 & 2.5 & 10.6 \\
\bottomrule
\end{tabular}
\label{tab:photoreal}
\end{table}

\subsection{On held-out frames, fewer frames cost 4 to 5\,dB and the person handling costs fidelity in observed pixels}
To measure rather than judge by eye, two zones were re-reconstructed with
every eighth posed frame held out (Table~\ref{tab:heldout}): 9 of 73 frames in
the racking aisle and 6 of 49 in the client market. Each run trained on the
remaining frames and rendered the held-out cameras, and PSNR, SSIM and LPIPS
were computed against the video frame inside the unmasked pixels only, about
half of each frame (52\% and 55\%), so that the presenters, whom the trainer is
meant to remove, are not scored. Three variants were compared per zone: the
trainer as used for the twin (all remaining frames, background fill and hull
carving); the same trainer given 8 frames evenly spaced among them, the
number the world model received for the racking aisle; and the trainer
without background fill and hull carving, as an ablation of the two changes
described above.

\begin{table}[t]
\caption{Held-out novel-view scores, mean over the held-out frames of each
zone, computed in unmasked pixels only. Higher PSNR and SSIM and lower LPIPS
are better. Each row is one training run of 3{,}000 steps at 480 pixels wide on
four CPU cores; per-frame scores are in the repository.}
\centering\small
\begin{tabular}{@{}llrrrrrr@{}}
\toprule
Zone & Variant & Train & Test & PSNR (dB) & SSIM & LPIPS & Splats \\
\midrule
Racking aisle & full trainer & 64 & 9 & 21.16 & 0.799 & 0.094 & 149{,}063 \\
 & 8 training frames & 8 & 9 & 17.34 & 0.632 & 0.144 & 142{,}124 \\
 & no fill, no carving & 64 & 9 & 23.64 & 0.893 & 0.041 & 189{,}057 \\
\midrule
Client market & full trainer & 43 & 6 & 23.10 & 0.815 & 0.127 & 121{,}818 \\
 & 8 training frames & 8 & 6 & 17.69 & 0.660 & 0.161 & 123{,}703 \\
 & no fill, no carving & 43 & 6 & 23.64 & 0.837 & 0.101 & 163{,}666 \\
\bottomrule
\end{tabular}
\label{tab:heldout}
\end{table}

Frame count is the largest effect. With 8 training frames instead of 64 or 43,
held-out PSNR fell by 3.8\,dB in the racking aisle and 5.4\,dB in the market,
and SSIM by 0.17 and 0.16. This is the regime the world model works in, and it
is where a learned prior has the most room to help; the 8-frame run is
therefore the matched-input baseline for the world-model capture, which could
not be scored on these frames itself because its estimated camera poses were
not stored and recovering them would have needed further paid calls. The
8-frame baseline is also optimistic: its camera poses and initial points come
from structure from motion over all frames of the zone, including the held-out
ones, which 8 frames alone would not reliably provide.

\begin{figure}[t]
\centering
\includegraphics[width=\linewidth]{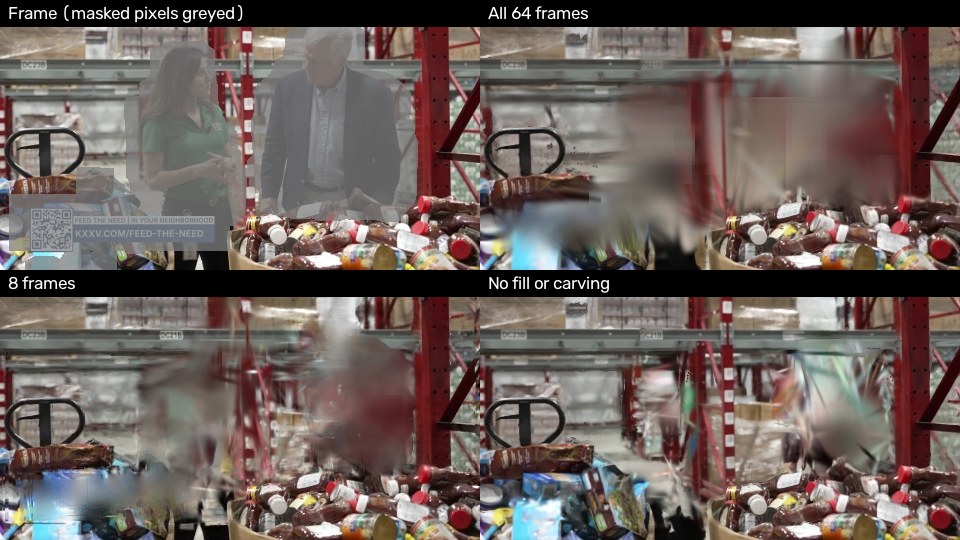}
\caption{Held-out frame f\_0184 of the racking aisle, which no run trained on.
Top left: the video frame, with the masked pixels that are not scored shown in
grey. Top right: the full trainer (PSNR 19.97\,dB in unmasked pixels). Bottom
left: the same trainer from 8 frames. Bottom right: without background fill
and hull carving, where the region the presenters covered holds streaks and
dark fragments instead of a soft blur. That region is not scored.}
\label{fig:heldout}
\end{figure}

The ablation went the other way from what the visual inspection suggested.
Without background fill and hull carving, held-out PSNR was
2.5\,dB higher in the racking aisle and 0.5\,dB in the
market (Table~\ref{tab:heldout}). The two changes turn the region where the presenters
stood into a soft blur rather than streaks and dark fragments
(Figure~\ref{fig:heldout}), but those regions are masked in the held-out frames too, so no ground truth exists
for them and their gain cannot be measured this way; what the metric does
show is the cost, in observed pixels, of training against a synthesized
background plate and of deleting splats that a few frames placed inside the
people's silhouettes. A fair score for the cleaned regions would need frames
of the same spot without the presenters, which this video does not contain.

To gauge run-to-run variation, the full-trainer runs were repeated with a
second random seed. Held-out PSNR changed by 0.13\,dB in the racking aisle
(21.16 to 21.29) and 0.05\,dB in the market (23.10 to 23.15), SSIM by at
most 0.008 and LPIPS by at most 0.007, so the frame-count and ablation
differences above are one to two orders of magnitude larger than the seed
spread. Two seeds give a range, not a confidence interval.

Three limitations of this protocol should be kept in view. LPIPS is computed on
the whole frame after the masked pixels are copied from the video into the
render, so it is diluted by roughly half and is comparable only between rows
of the same zone. The runs use 3{,}000 steps at 480 pixels, half the steps of
the documented zones, to fit the evaluation on a CPU. And the structure from
motion that initializes every run saw the held-out frames, so the scores
measure the splat optimization, not the pose estimation.

\subsection{World-model output becomes a valid, metric OpenUSD asset}
Converting the splat world to OpenUSD required four corrections, described in
Materials and Methods: a coordinate transform from the world model's camera
frame, an untyped reference prim so the splat schema is not hidden, binary
packaging with correct alignment, and a choice of splat density by target.
After conversion the pipeline re-opens the package and fails unless it finds
a \code{ParticleField3DGaussianSplat} prim with positions. In a synthetic test
with 20{,}000 splats at spherical-harmonic degree 1, the SPZ file was 262\,KB
and the USDZ package 2.0\,MB, a 7.7-fold increase, because SPZ is quantized
and compressed while USD stores 32-bit floats and USDZ is uncompressed by
specification. For the racking aisle, the 270{,}423-splat SPZ file (spherical
harmonics of degree 0) is 2.1\,MB and its USDZ package 17.5\,MB, an 8.5-fold
increase. The three zones are composed into one stage, \code{/World/Site}, in
which each zone is a separately referenced layer with its own placement, so a
zone can be re-captured without touching the others; the packaged site is
48.5\,MB for about 747{,}000 splats.

\subsection{The same artifacts drive a browser, Blender and Omniverse}
The reconstruction step's output is consumed unchanged by the rest of the
system, which we summarize here and detail in Appendix~\ref{app:system}. A
serverless backend on AWS serves simulated cold-chain telemetry and an
operations model in the twin's metric frame; a facility model scripted in
Blender supplies the building around the captured zones; a browser viewer
renders the captures with Spark \cite{spark}; and NVIDIA Omniverse rendered the
modeled facility on a rented GPU but drew the splat prims blank grey
(Appendix~\ref{app:system}). None of these parts learns anything; they are
included to show that a world-model capture can be used as an ordinary,
metric OpenUSD asset by other tools.

\subsection{Summary of measurements}
Table~\ref{tab:results} collects the cost, time and fidelity measurements, the
three properties that decide whether a small operator can adopt the approach.

\begin{table}[t]
\caption{Measurements for the twin. All values are single measurements; fusion time was not recorded. Held-out novel-view scores are in Table~\ref{tab:heldout}.}
\centering\small
\begin{tabular}{@{}lll@{}}
\toprule
Quantity (unit) & Value & How measured \\
\midrule
Input frames used & 20 (8 + 6 + 6) & hand-picked from 40 per zone \\
Frame extraction, racking window (s) & 88 & wall clock, 4 CPU cores \\
World-model cost, posing (credits; USD) & 200; 0.20 & 10 credits per frame, rate card \\
World-model cost, failed generation (credits) & 0 to 800 & billing of failures unknown \\
Splats, racking / sorting / market & 270k / 282k / 195k & fused files \\
SPZ size, racking / sorting / market (MB) & 2.1 / 2.2 / 1.6 & committed files \\
USDZ size, racking / site (MB) & 17.5 / 48.5 & committed files \\
World-model generation time (min) & n/a & both runs failed \\
SPZ to USDZ, 270k / 182k splats (s) & 5.2 / 3.9 & wall clock, 4 cores \\
Locally trained zones (credits) & 0 & four zones, Table~\ref{tab:photoreal} \\
Modeled facility, GLB / USDZ (MB) & 1.8 / 1.6 & committed files \\
Telemetry latency, p50 / p95 (ms) & 67 / 78 & cloud client, 1{,}000 requests \\
AWS cost, October 2026 to date (USD) & 0.00 & Cost Explorer, demo traffic \\
Browser download, default / captures (MB) & 2.7 / 60 & headless Chromium \\
Browser frame rate, no GPU (fps) & 1.1 / 0.5 & software WebGL, a floor \\
Omniverse offline render, 96 frames (s) & 18.6 to 49.4 & A10G, 720p, 8 subframes \\
Omniverse GPU trial, 7 runs (min; USD) & 94; 1.34 & list price, terminated \\
\bottomrule
\end{tabular}
\label{tab:results}
\end{table}

\section{Discussion}
This study set out to learn what current world models and Gaussian splatting
recover from in-the-wild video. The main finding is that the observed parts of
a real scene come back cheaply and at metric scale, while unobserved space does
not yet. A consequence is that a spatial digital twin does not have to start
with a CAD model or a laser scanner. Every input to this one was publicly available: a broadcast news
tour, published facility figures and open-source or pay-per-use tools. No
drawing, scan or site visit was used. A short walkthrough video, a world model
and OpenUSD's native splat schema produce a metric twin, Blender supplies the parts the camera never saw, and a serverless data
layer connects it to the systems an operator already has for almost no
standing cost. One author, working with AI coding assistants, went from the
first commit to a deployed backend in seven days (19 to 25 October 2026). That
speed comes from composing existing technologies through open interfaces,
not from new models. Keeping OpenUSD and a small JSON contract as the only interfaces makes
each step replaceable as world models, data platforms and renderers improve,
and lets an organization choose a renderer by hardware, cost and license
posture: OpenUSD is open source, Blender is GPL, and Omniverse combines open
components with NVIDIA-licensed runtime pieces. On a public interview video, 20 frames and
200 credits of world-model calls produced a three-zone OpenUSD twin that the
browser renders with live telemetry.

The main limitation in the current results is coverage. Twenty frames from an
interview walkthrough recover the parts of three zones the camera pointed at,
not whole rooms, and the generative step that should fill the rest failed on
the provider's side. Until it works, the whole-building view is a designed
model, which explains the operation well but is not a reconstruction; the
honest summary is that video and world models gave the twin its photographs
of real places, while its map is drawn. A second limitation is that generated is not measured. A world model completes
what the camera did not see. That is useful for a planning or communication
twin and unsafe if someone reads a generated aisle as the real one. Generated
regions should be marked in the stage, for example with an attribute on splats
or a separate USD layer, so renderers can show them differently.
No generated regions exist in the twin yet, so no such marking has been
implemented; designed content is kept in its own model and layer so it cannot
be mistaken for a capture. Scale comes from the world model's metric
estimate rather than a survey, so absolute dimensions should be checked
against at least one known measurement before the twin informs layout
decisions. Held-out scores exist only for the locally trained zones: the
world-model capture could not be scored on the same frames because its
camera poses were not stored, so the comparison between the two rests on the
8-frame local baseline and on inspection by eye. People, motion blur, reflective steel and repetitive racking remain
hard inputs; masking people before reconstruction would help, and the world
model's API exposes masked inputs that we have not yet evaluated.

The twin is also a composite by design. Its captured geometry comes from a
video that appears to show the food bank's Waco hub, its operating volumes
come from published figures for the 135{,}000\,sq\,ft Austin facility, the
rest of the building is designed, and the data is simulated. That makes it a
fair picture of what a Texas food bank's twin looks like, but not a record of
any one site, and its provenance (Table~\ref{tab:provenance}) should travel
with every image of it. A twin of a real facility would replace each designed
or simulated row with a capture or a live feed from that facility. Keeping these sources in one frame is what lets the twin be
used at all: one USD stage composes the modeled facility, the captured zones
(by reference, placed by the same configuration as the browser) and the
operations layer, and cropping is a non-destructive opacity override, so the
capture library is never copied. A new capture or a new data feed therefore changes one layer rather
than the whole stage.

Applying the approach also needs care outside the software. Walkthrough
videos capture staff, volunteers and clients, so faces should be removed or the
space filmed when empty, and a facility's permission obtained before a
reconstruction is published. The next steps are a first real data connector,
most cheaply temperature loggers publishing over MQTT to AWS IoT Core
\cite{awsiotcore}, a person-aware frame picker, and training the locally reconstructed zones at full resolution on a GPU. The
comparison with a conventional splat reconstruction shows the division of
labor: the world model gives metric scale and usable geometry from eight
frames, while local training needs a scale assumption and, on held-out frames,
loses 4 to 5\,dB of PSNR when cut from 43 or 64 frames to 8
(Table~\ref{tab:heldout}). Few-frame reconstruction is where a learned prior
has the most to add, and scoring a world model on the same held-out frames is
the most direct next experiment.

\section{Materials and Methods}

\subsection{Experimental Design}
The study asks what a world model and locally optimized Gaussian splats
recover from one uncurated public video, and what that costs
(Table~\ref{tab:results}); the twin is the test case that makes the
question concrete.
The design was fixed before any paid call: one zone first (the racking aisle,
eight frames), then two more zones with six frames each, the lowest-cost API
tasks available, and a project ceiling of 1{,}000 credits. All code, the AWS template and the input
frames are in the project repository.

\subsection{Frame extraction}
World-model APIs accept only a few images (Marble's Auto Layout multi-image mode
accepts up to eight \cite{worldlabs_multiimage}). The ingest
stage samples a window of the video and, for every interval of $N$ seconds,
scores candidate frames with ffmpeg's blur detector and keeps the sharpest;
interlaced frames are deinterlaced only when flagged. A hard cap of 40 frames
bounds storage and upload. For the case study, frames were taken from
5:05--7:45 at one per 4\,s with the bottom 264 pixels cropped to remove the
broadcast graphic, giving 40 candidates at 1920\,$\times$\,816 in 88\,s of wall-clock time on
four CPU cores for the 160\,s window; all inputs
share one resolution, which reconstruction requires. The automatic picker
splits the pool into eight equal segments and keeps the frame with the highest
Laplacian variance in each; for the case study it was replaced by hand
selection as described in Results. A text prompt described the space and asked
for no people.

\subsection{World-model reconstruction}
The pipeline implements two routes on the World Labs platform. The first, a
multi-image request to Marble that returns a finished Gaussian-splat world in
SPZ format \cite{spz} with a metric scale factor and ground-plane offset
\cite{worldlabs_spz}, needs a Marble v1 key and was not available to the
account used here. The reported twin was built with the second route, through
the Marble 2 developer API, in three stages. In the pose stage,
\code{images2PosedRGBD} returns a camera pose, metric depth and confidence for
each input frame (10 credits per frame on the rate card of 19 October 2026,
where 1{,}000 credits cost US\$1). In the generate stage, \code{atlasGenerate}
is asked for new posed RGB-D views from target cameras around the zone, with
the masked presenters excluded from its context (50 credits per view); this is
the step in which the world model extrapolates unseen space, and it failed in
both attempts (Results). In the free, local fuse stage, every RGB-D view is
unprojected, the most confident sample per 1\,cm voxel is kept, masked pixels
are dropped, and the points are written as isotropic Gaussians with spherical
harmonics of degree 0, as PLY and SPZ version 2. The floor is taken, in order
of preference, from a dense 5\,cm horizontal slab of points, from the world
model's ground plane, or from a fixed 1.5\,m below the cameras. A site stage
then places each zone on a plan in meters (racking at the origin, sorting at
$+26$\,m and the market at $-24$\,m on $x$) and writes one USD stage that
references each zone's layer. This plan is a library layout, not a survey;
the composed twin re-places the zones as described in Appendix~\ref{app:system}.

\subsection{Locally trained zones}
The photoreal route uses no paid service. Frames from a zone's segment of the
video are extracted at 4 to 6 frames per second and 960 pixels wide. People are
masked with a DeepLab v3 segmentation model run through MediaPipe, the masks
are grown by 35 pixels and unioned with neighboring frames, and the broadcast
graphic is masked by a fixed box. Under the masks, a background plate is built
from pixels warped by homography from frames up to 15\,s away in which the
presenters had moved, with a blurred inpaint where no frame saw the
background. Frames are kept if they are not in the
blurriest 20\%, at least 35\% of their pixels are unmasked, and the median
optical flow since the last kept frame is at least 14 pixels. COLMAP 3.9
estimates poses with masked SIFT features on the CPU, with sequential matching,
or exhaustive matching and a 3$^\circ$ initial triangulation angle for the
low-texture sorting room. The trainer implements
3D Gaussian Splatting \cite{kerbl20233d} in plain PyTorch with a sparse
per-pixel rasterizer, so that it runs without a GPU (about 1\,s per step at
480 pixels on four cores): elliptical weighted average projection, front-to-back
compositing, an L1 plus 0.2 D-SSIM loss that sees the background plate at 0.3
weight under the masks, visual-hull carving every 250 steps (a splat is removed
when its center falls on a presenter mask in more than half of the five or
more frames that see it), adaptive cloning, splitting and pruning, and opacity
resets every 3{,}000 steps, at half resolution for the first quarter of the
steps. The documented run trains each zone for 6{,}000 steps, roughly 1.7\,h
at the rate above; per-zone wall-clock times were not recorded. Export levels
the scene to the mean camera up direction, takes the floor from the lowest
well-supported band of structure-from-motion points below the cameras, scales
so that the cameras sit 1.6\,m above it, keeps splats inside at least six
camera frustums and within 80\,m of the path, and writes the same SPZ, USDZ
and index files as the world-model zones; the composed twin crops each zone to
its room.

\subsection{Held-out evaluation}
For the racking aisle and the client market, frames were extracted and masked
as above and posed with COLMAP 3.9.1, which registered 73 and 49 frames. The
posed frames were sorted by name and every eighth one (index $i \bmod 8 = 4$)
was held out: f\_0121, f\_0132, f\_0184, f\_0239, f\_0247, f\_0272,
f\_0283, f\_0304 and f\_0325 for the racking aisle, and f\_0020, f\_0030,
f\_0038, f\_0067, f\_0079 and f\_0091 for the market. The 8-frame variant
keeps 8 of the remaining frames at evenly spaced indices. Every run trains for
3{,}000 steps at 480 pixels wide with seed 0, at most 48 splats per pixel and
200{,}000 splats in total, on four CPU cores; the ablation sets the background
plate weight to 0 and disables hull carving. Training took 53 to 96\,min per
run. The held-out cameras are rendered from the trained splats at the same
resolution. PSNR and SSIM are computed over the unmasked pixels, using the
person and graphic masks without their background plate; LPIPS (AlexNet) is
computed on the full frame after the masked pixels of the render are replaced
by the video frame's.

\subsection{Conversion to OpenUSD}
The splat world is converted with \code{usd-convert-gsplat} 0.1.15
\cite{usdconvertgsplat} into a
\code{ParticleField3DGaussianSplat} prim.
World Labs SPZ files use an OpenCV-style camera convention
(\code{marble\_raw\_opencv}, Y down) in raw units that the world's metric scale
factor converts to meters \cite{worldlabs_spz}. Leaving the splat payload untouched, we place it
under a parent transform, as shown in Equation~\ref{eq:xform},
\begin{equation} \label{eq:xform}
  M = R_x(180^\circ)\, T(0, -o, 0)\, S(s),
\end{equation}
where $s$ is the world's metric scale factor and $o$ its ground-plane offset,
and set the stage to Y-up with \code{metersPerUnit}\,=\,1. With synthetic
input the ground lands at exactly $y=0$. A typed \code{Xform} that references
the splat would override the referenced prim's schema type, so the pipeline
authors \code{/World/FoodBank} as the transforming \code{Xform} and
\code{/World/FoodBank/Splat} as an untyped prim carrying the reference. The
converter's direct USDZ output is an ASCII layer zipped without the 64-byte
alignment the USDZ specification requires, so we convert to binary
\code{.usdc} and package with OpenUSD's \code{UsdUtils.CreateNewUsdzPackage}.
Conversion took 5.2\,s for the 270{,}423-splat world-model racking zone and
3.9\,s for a 181{,}556-splat first version of the locally trained one on four
CPU cores. The schema
family defines half-precision companions to its attributes, such as
\code{positionsh} \cite{usd_particle_position}, that would roughly halve file
size; the converter does not yet write them.

\subsection{Statistical Analysis}
This is a single case study, and every value reported is a single measurement unless stated:
counts and file sizes from the committed files, times from one wall-clock run on
a four-core CPU machine, and costs from rate cards or list prices. No repeated
runs are reported except where stated. The weekday ranges
of the operations model are the minimum, median and maximum over 40 simulated
weekdays from one deterministic run. The held-out evaluation
(Table~\ref{tab:heldout}) reports the mean over 9 or 6 held-out frames of one
training run per row, with the two full-trainer rows repeated with a second
seed; no hypothesis tests or confidence intervals are
computed, and differences of a few tenths of a decibel should not be read as
effects. Telemetry latency is the median and 95th percentile of 1{,}000
client-side requests (Appendix~\ref{app:system}).

\subsection{Animal and Human Studies}
No human subjects research was performed. The input is a publicly broadcast
news video; people who appear in it were excluded from input frames where
possible and the world model was prompted to omit them. The video is used as
published by its broadcaster; the twin is a representative facility inspired
by it, not a record of the building.

\section*{Acknowledgments}

\subsection*{General}
This project was inspired by 25 News KXXV's public video tour of the Central
Texas Food Bank's new facility \cite{kxxv2026ctfb,kxxv2026waco} and by the
food bank's published account of its work. Neither organization was involved
in or endorsed this study. During the preparation of this work, the author used Claude
(Anthropic) in order to write and test the pipeline software and to draft the
manuscript. After using this tool, the author reviewed and edited the content
as needed and takes full responsibility for the content of the publication.

\subsection*{Author Contributions}
S. Mohapatra conceived the idea, designed the study, conducted the
experiments and wrote the manuscript.

\subsection*{Funding}
This research received no specific funding.

\subsection*{Competing interests}
The author declares that there is no conflict of interest regarding the
publication of this article.

\subsection*{Data Availability}
Code, the AWS template, the operations master data and the case-study input
frames are available at
\url{https://github.com/pikulsomesh/foodbank-spatial-twin}. The source video
is publicly available from its publisher \cite{kxxv2026ctfb}. The fused zone
files (SPZ), the composed USD site and the per-zone placements are in the same
repository. The code is released under the MIT License. The held-out
evaluation is reproduced by \code{scripts/splat\_train.py --holdout 8} and
\code{scripts/splat\_eval.py}, and its per-frame scores are in
\code{docs/eval/}.

\section*{Supplementary Materials}
Movies S1 to S3 are available at
\url{https://pikulsomesh.github.io/foodbank-spatial-twin/paper/}
(\code{movie\_S1.mp4}, \code{movie\_S2.mp4} and \code{movie\_S3.mp4}).

Movie S1. A 22\,s overview of the pipeline, 1080\,$\times$\,1350: the public
news tour that inspired the twin, a world-model splat capture, the pipeline
diagram, three NVIDIA Omniverse RTX shots of the modeled facility (Movie~S3),
the browser twin with simulated telemetry, and an end card. Omniverse shots are
labeled as such.

Movie S2. A 24\,s 1080p flythrough of the OpenUSD stage, rendered with the CPU
Gaussian-splat renderer.

Movie S3. Three 4\,s NVIDIA Omniverse RTX clips of the modeled facility
(facility orbit, racking aisle, forklift chase), 1280\,$\times$\,720 at 24 frames per
second, rendered headless on an EC2 g5.xlarge; the forklift follows a scripted
path.

Data file S1. Commands to reproduce the pipeline end to end, in three files in
the repository's \code{docs} folder: \code{CTFB\_WAREHOUSE\_RUN.md}
(world-model zones), \code{PHOTOREAL.md} (locally trained zones) and
\code{RENDERING.md} (CPU, Blender and Omniverse renders), with
\code{scripts/build\_facility.py} for the modeled facility.

Data file S2. The architecture diagrams as editable draw.io files
(\code{paper/diagrams}).

\appendix
\section{System details}
\label{app:system}
This appendix holds the system parts of the study that do not bear on the learning questions: the data backend, the operations model, the modeled facility, the viewers and renderers, cost and operational technology security.

\subsection{Results}
\subsubsection{Live data reaches the twin at near-zero standing cost}
\begin{sidewaysfigure}
\centering
\includegraphics[width=\linewidth]{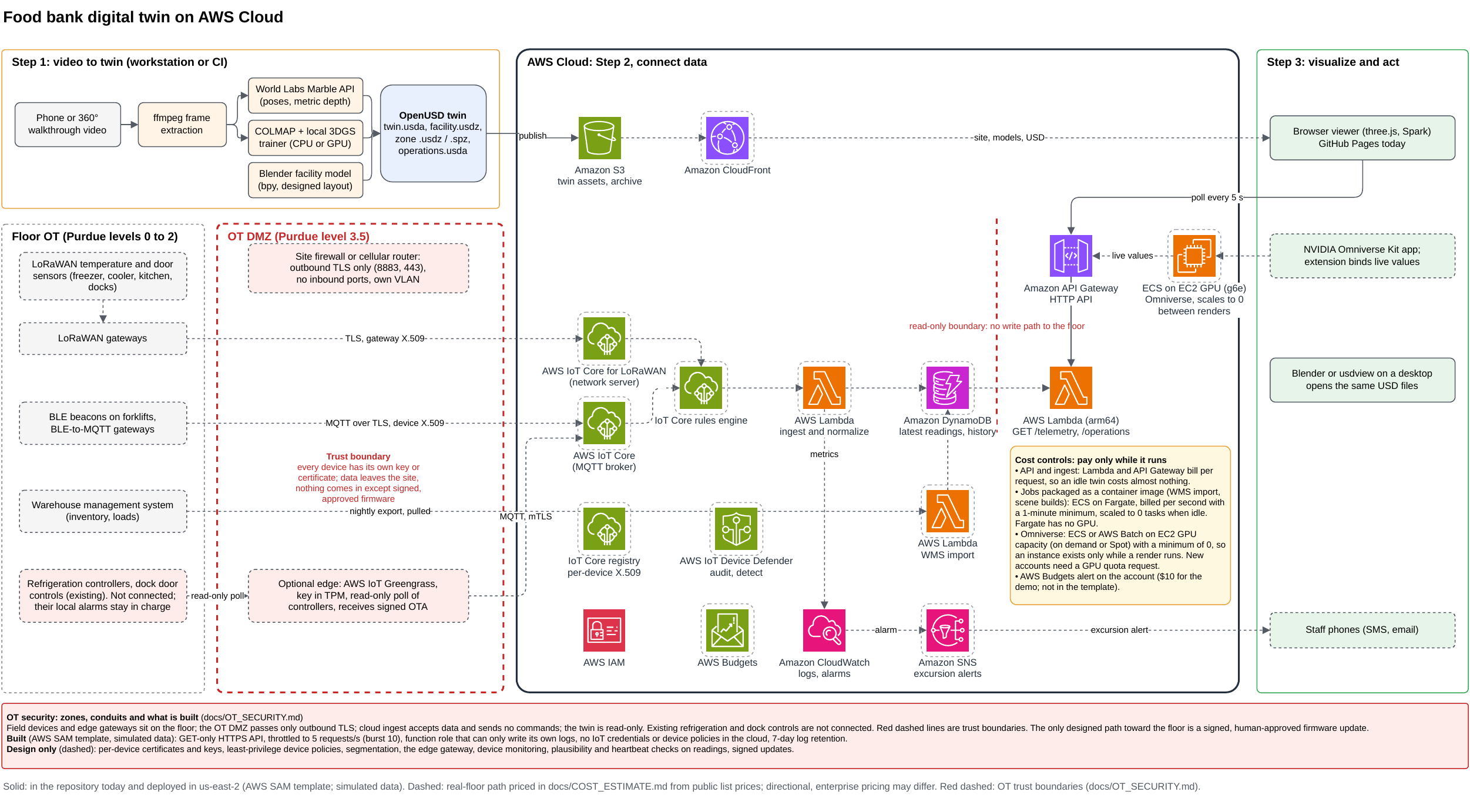}
\caption{The food bank twin on AWS. Solid borders are in the repository today
(AWS Serverless Application Model template; simulated data); dashed borders are
the real-floor path priced in Table~\ref{tab:costest}. Step~1 runs on a
workstation or in continuous integration and publishes the OpenUSD twin to
object storage; Step~2 connects floor sensors and the warehouse management
system; Step~3 is the browser, Omniverse on an on-demand or Spot GPU
instance that is stopped after each session, desktop tools and staff alerts.
The cost-controls panel lists how each part bills only while it runs. Red
dashed lines are OT trust boundaries (Purdue levels 0 to 2 on the floor and an
OT DMZ at level 3.5); the twin has no write path to the floor, and the bottom
panel separates the controls that are built from those that are design only. Drawn in draw.io with the AWS architecture
icons; the source is in the repository's \code{paper/diagrams} folder.}
\label{fig:aws}
\end{sidewaysfigure}

The reference backend serves simulated cold-chain telemetry for three zones
(freezer, cooler and dry storage, with setpoints of $-18^\circ$C, $3^\circ$C
and $21^\circ$C) and one forklift on a closed path from the dock through the
storage aisles. It has no virtual private cloud, NAT gateway, database or
provisioned capacity, so an idle twin costs nothing beyond storage.
The same function also serves \code{GET /operations}, the operations layer
described next. No real data connector is deployed yet, so all telemetry in
this paper is simulated. Figure~\ref{fig:aws} shows the AWS pattern: the
services in the repository's template today (object storage, the HTTP API and
its function, logs and access roles) and the real-floor path
that this appendix prices, in which sensors and the warehouse management
system feed the same API. Appendix~\ref{app:clouds} gives equivalent designs
on Microsoft Azure and Google Cloud. The stack was deployed on 25 October 2026
to us-east-2, because the account's organization policy blocks the function,
API and deployment services in us-east-1. AWS Cost Explorer showed \$0.00
billed for October 2026 at that date: at demo traffic the function, HTTP API and
bucket stay within the free tier. Over 1{,}000 sequential requests to
\code{GET /telemetry} from a cloud container in the United States, paced at
about three per second, all 1{,}000 succeeded, with a median round trip of
67\,ms and a 95th percentile of 78\,ms. In an earlier run of the same size,
1.4\% of requests were refused by the API's deliberate throttle (HTTP 429).
These are measurements from one individual AWS account under its own quotas
and policies, not a user's device or an enterprise account.

\subsubsection{An operations layer turns the geometry into a working food bank}
\begin{figure}[t]
\centering
\includegraphics[width=0.85\linewidth]{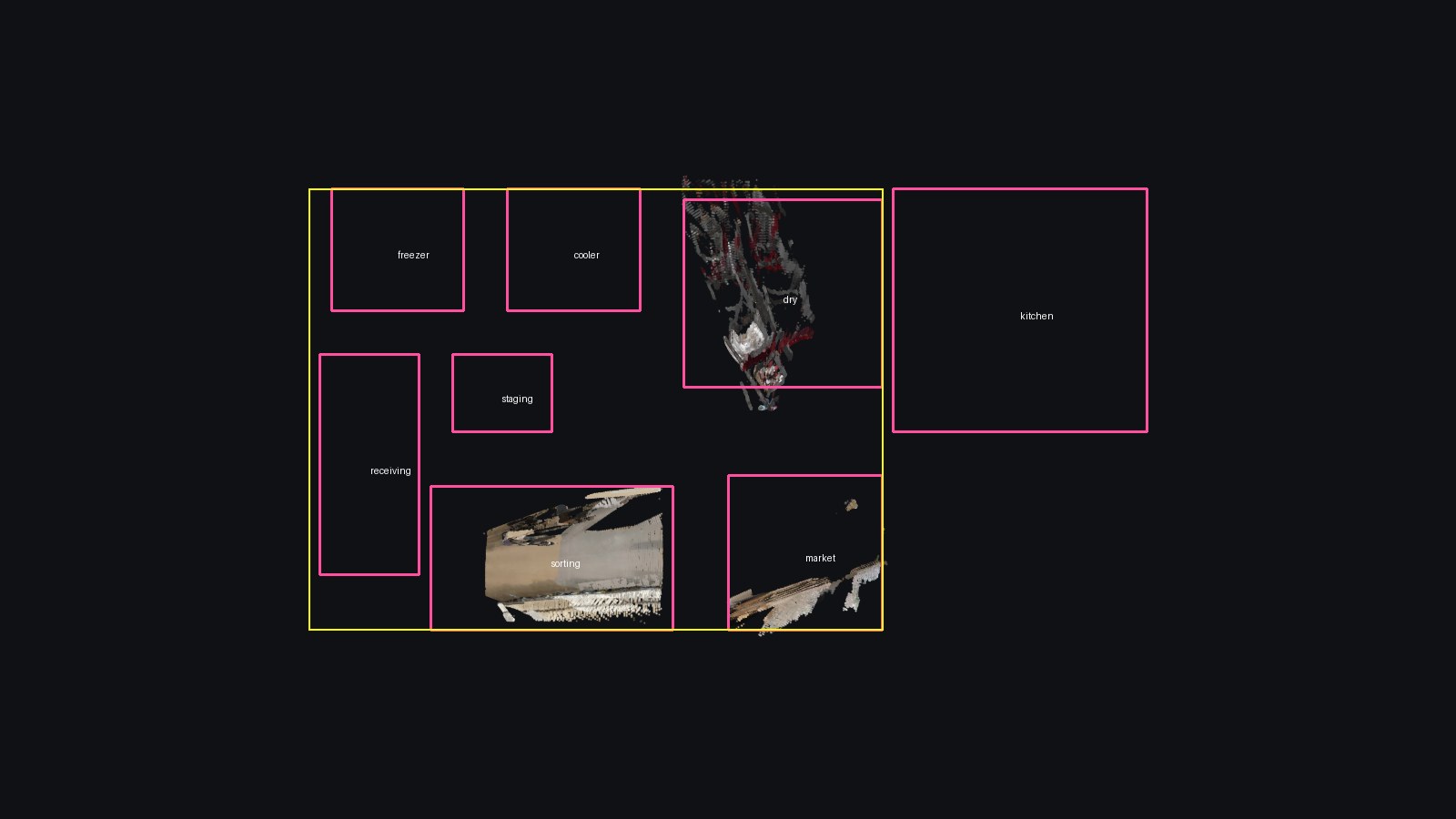}
\caption{Alignment check: top view of an earlier version of the composed
OpenUSD stage, rendered on the CPU, with the operations-layer footprints drawn
over it (pink: areas; yellow: building). The world-model captures of the
sorting room, dry racking and market land in their areas; the racking capture
extends slightly past its footprint. The current stage keeps these placements
and adds the trained zones and the modeled facility.}
\label{fig:alignment}
\end{figure}

The captured zones show what the building looks like; an operations layer
models what it does, and it follows the same three steps. In the twin's metric
frame, eight operational areas (receiving, product recovery or sorting,
freezer, cooler, dry racking, order staging, the client market and the
community kitchen) are footprints tagged with the capture that covers them. On
the data side, the same stateless function that serves telemetry also answers
\code{GET /operations} with inventory by product family, flows between
sources, areas and outbound channels, dock schedules and kitchen batches, all
computed from the clock and one master-data file, with no database; a
JavaScript port returns identical payloads offline, which a test enforces. For
Omniverse, a script writes the layer as \code{operations.usda}, with one prim
per area (8), kitchen station (7) and flow (16), keyed by the same identifiers
as the API so that an extension can bind live values to them. One stage,
\code{twin.usda}, composes the twin for Omniverse: the modeled facility
described below (visible), the four captured zones (the locally trained
versions, falling back to the world-model capture where an area has none;
authored invisible so they can be switched on), and this layer. Each capture
is referenced, not copied, and placed with the same configuration file the
browser uses, so Omniverse, the browser and the API share one frame
(Fig.~\ref{fig:alignment}). Each capture is cropped to its area by a 0.65\,MB
sublayer that only overrides the opacities of splats outside the crop box, so
the capture library is never copied or edited, and a test fails if the
placements in the stage and in the browser configuration drift apart. Opened
with the OpenUSD library, the stage holds 452 prims, including 137 meshes and
four Gaussian-splat prims, in Y-up meters.

The kitchen is modeled as a small factory, with a bill of materials and a
routing per recipe: ingredients are picked from storage, and each batch of 250
meals moves through prep, cooking on a kettle, tilt skillet or combi oven,
blast chilling, portioning and packing, and cold holding until dispatch. The
same stance extends to the whole distribution center: inbound loads from five
source types either pass through the sorting room or go straight to put-away,
stock waits in its temperature zone with a days-of-supply target, and outbound
pounds leave for partner agencies, mobile pantries, the market and the kitchen.
Volumes are anchored to published figures for the Central Texas Food Bank's
main facility (Table~\ref{tab:published}); every other parameter, including
source and channel shares, product mixes, pallet weights, capacities, time
windows, equipment counts, recipes and step times, is an estimate and is
flagged as one in the master data.

Over eight simulated weeks, a weekday received 231{,}000 to 296{,}000\,lb
(median 258{,}000) and distributed 220{,}000 to 300{,}000\,lb (median 260{,}000,
about 217{,}000 meal equivalents), with 4.49 to 4.63 million lb on hand, or
22.1 to 22.8 days of supply. Each weekday had 13 to 17 inbound trucks and 59 to
80 partner pickups, and the kitchen produced 912 to 997 meals; Saturdays run
at reduced volume and Sundays are closed. These numbers follow from the
published annual volume and the estimated parameters; they are a plausible
operating rhythm for the twin to display, not a measurement of the food bank.

\begin{table}[t]
\caption{Published figures that anchor the operations model. All other parameters are estimates.}
\centering\small
\begin{tabularx}{\linewidth}{@{}>{\raggedright\arraybackslash}Xl@{}}
\toprule
Figure & Source \\
\midrule
74 million lb distributed per year, 26.6 million lb of it produce (fiscal year 2024) & \cite{ctfb_impact_fy24} \\
251 partner agencies in a 21-county service area & \cite{ctfb_impact_fy24} \\
135{,}000\,sq\,ft facility & \cite{ctfb2016facility} \\
More than 7{,}000 pallet positions, more than 35{,}000\,sq\,ft of cooler and freezer, 15 docks & \cite{cfs_processing} \\
4{,}200\,sq\,ft production kitchen with kettles, combi ovens, blast chillers and tilt skillets & \cite{ctfb_kitchen_volunteer} \\
About 950 kitchen meals per day for after-school meals (2018); about 2{,}000 a day by 2025, including summer meal sites & \cite{ctfb_kids_cafe,ctfb_kitchen_history} \\
1.2\,lb of food per meal & \cite{fa_meal} \\
\bottomrule
\end{tabularx}
\label{tab:published}
\end{table}

\subsubsection{A modeled facility gives the whole-building view the captures cannot}
\begin{figure}[t]
\centering
\begin{subfigure}{0.62\textwidth}
  \includegraphics[width=\linewidth]{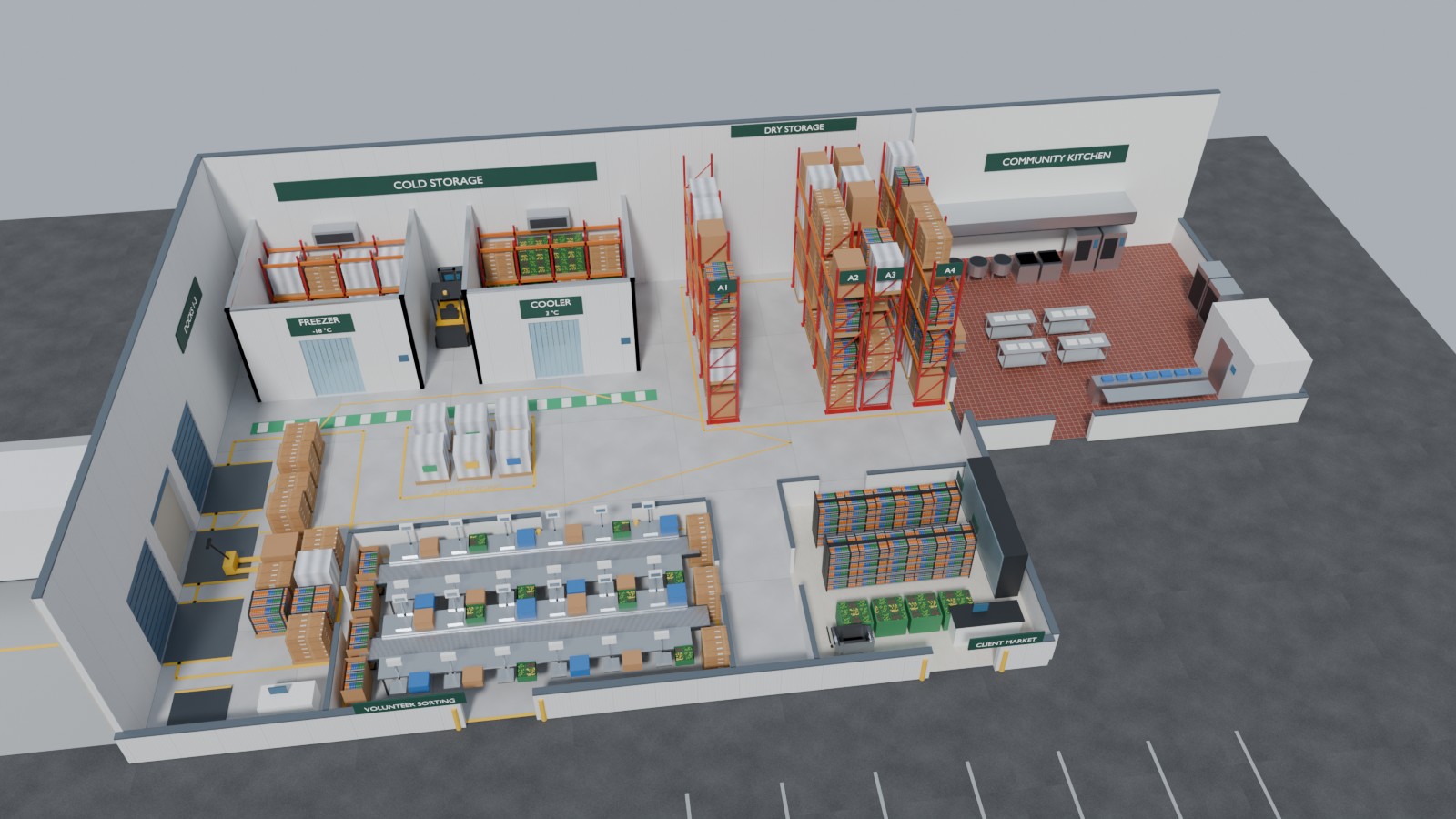}
  \caption{\label{fig:facilityA}}
\end{subfigure}\hfill
\begin{subfigure}{0.36\textwidth}
  \includegraphics[width=\linewidth]{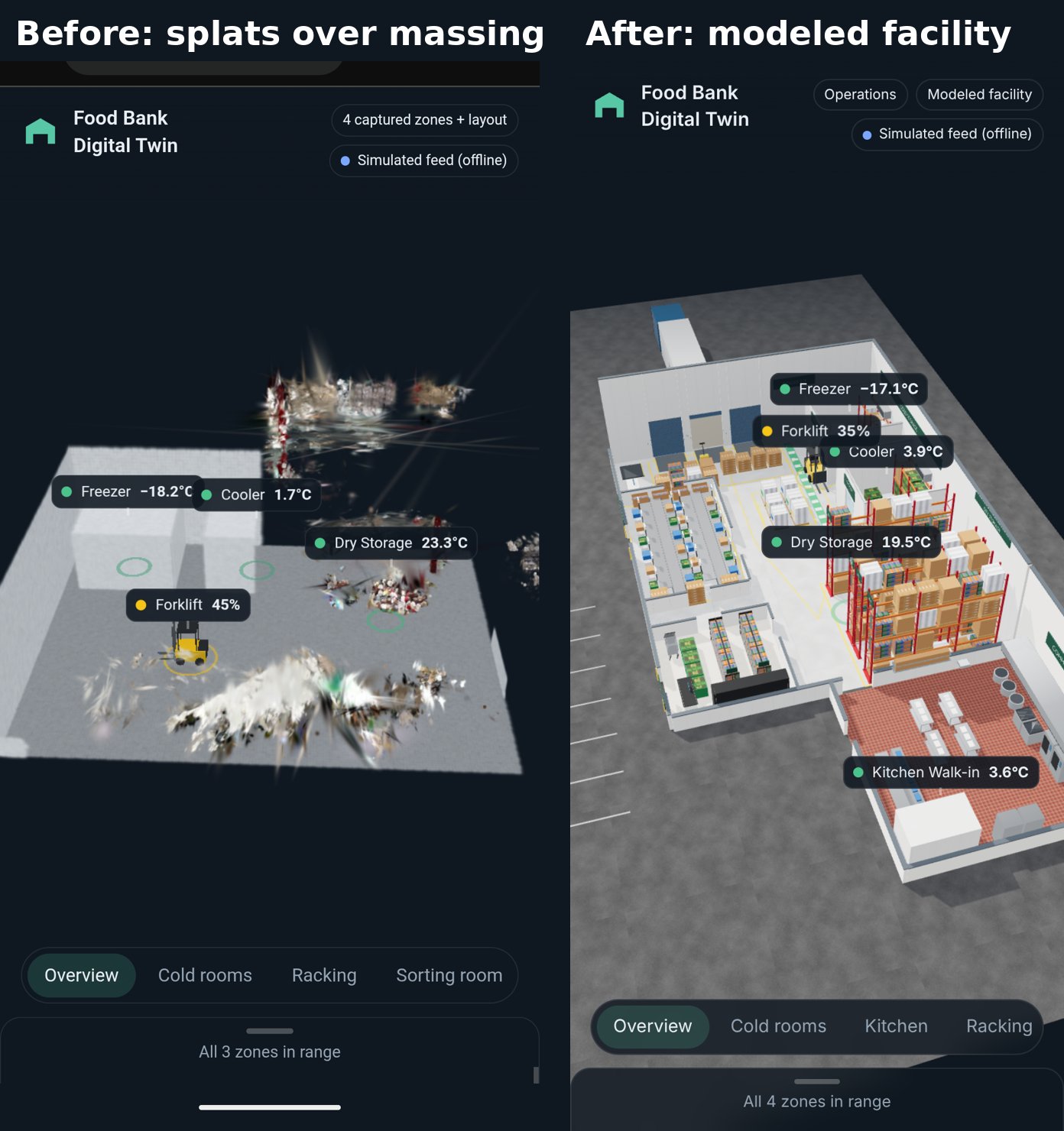}
  \caption{\label{fig:facilityB}}
\end{subfigure}
\caption{The modeled facility. (\subref{fig:facilityA}) Cycles render of the
Blender scene: cold storage, pallet racking, receiving and staging, the
volunteer sorting room, the client market and the community kitchen annex.
(\subref{fig:facilityB}) The browser overview on a phone before (captured
splats over a massing model) and after (the modeled facility), both with
simulated telemetry.}
\label{fig:facility}
\end{figure}

The captures answer \emph{what does this aisle look like}, but only from near
the camera path: from an overview they break into fragments, and a phone
screen showed the twin as debris over a grey massing model
(Figure~\ref{fig:facilityB}). For the view a manager or volunteer needs first,
where everything is and how it is doing, the twin's default scene is therefore
a modeled facility (Figure~\ref{fig:facilityA}). One script builds it headless
with Blender's Python module \cite{blender} in the telemetry frame, from the
operations layer's footprints and the cold-room probe positions, so the
labels, the forklift route and the operations page line up with it. It
contains the shell and dock doors with a trailer, the freezer and cooler as
cutaways with racking inside, four rows of selective pallet racking, receiving
and order staging, the volunteer sorting room with tables, roller conveyors and
scales, the client market with wire shelving, a produce display, a fridge case
and checkout, the community kitchen annex with kettles, tilt skillets, combi
ovens, blast chillers, prep tables, a pack line and a walk-in, forklifts,
painted lanes and signage. The same scene is written as a 1.8\,MB binary glTF
file for the browser and a 1.6\,MB USDZ package for Omniverse, which the
OpenUSD library opens as 129 meshes in Y-up meters. The model is designed, not
reconstructed: the fit-out is inspired by the walkthrough video and the layout
comes from the operations layer, the building is a compact 26\,$\times$\,20\,m
stand-in plus an annex rather than a 135{,}000\,sq\,ft warehouse, and the
viewer labels it a representative food bank, not a replica. The captures stay
one switch away, in the browser and in the USD stage.

\subsubsection{The browser shows the twin with live telemetry}
\begin{figure}[t]
\centering
\includegraphics[width=\linewidth]{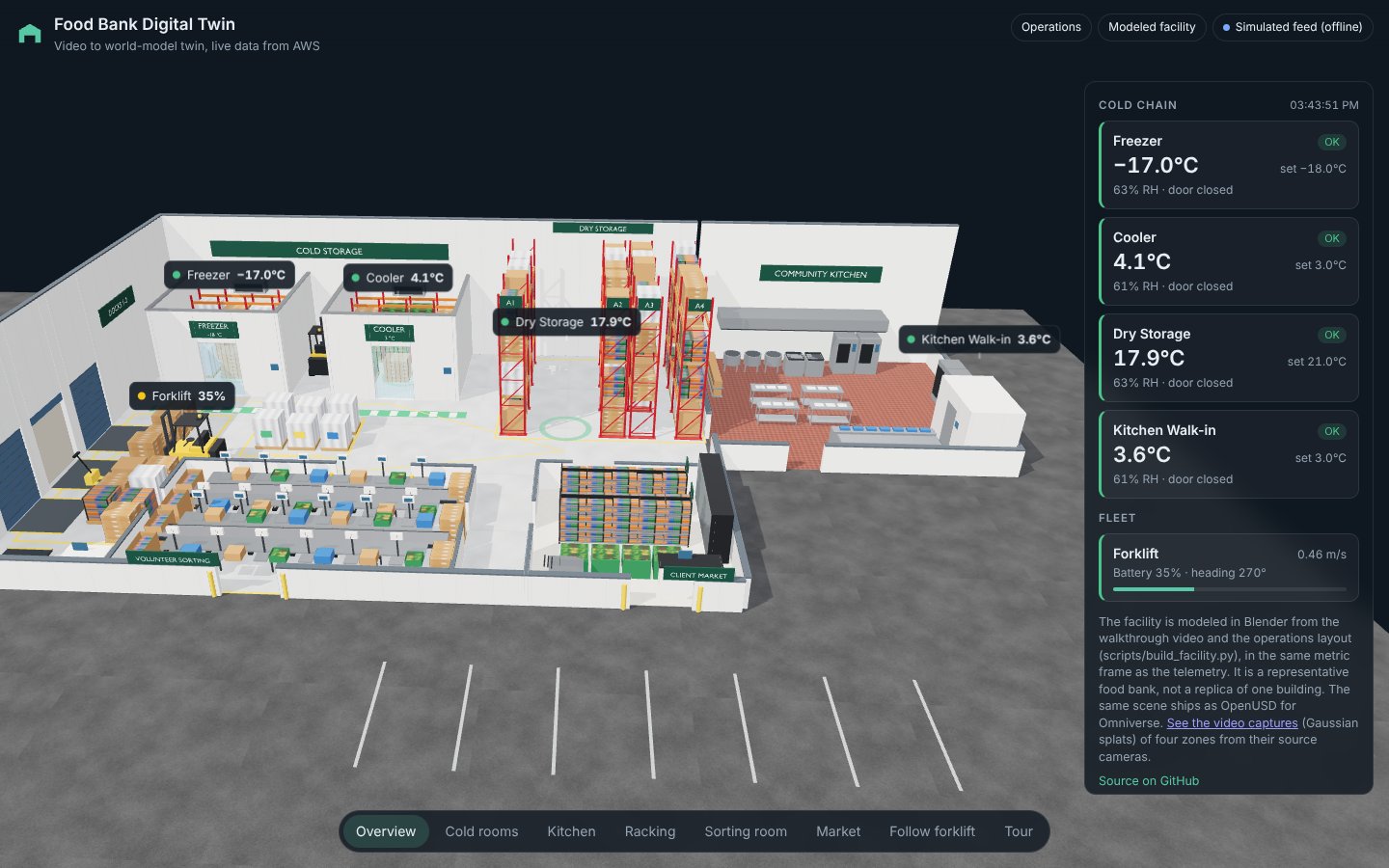}
\caption{The browser viewer on a desktop, opening on the modeled facility with
simulated telemetry labels and the cold-chain panel. The buttons open preset
views; the panel links to the captured zones.}
\label{fig:browser}
\end{figure}

The public demo at \url{https://pikulsomesh.github.io/foodbank-spatial-twin/}
opens on the modeled facility, drawn with three.js, with labels for the
freezer, cooler, dry storage and kitchen walk-in temperatures, the forklift's
battery, and a cold-chain panel (Figure~\ref{fig:browser}). The model's
forklift moves along the position and heading the telemetry reports. Preset
views open the cold rooms, kitchen, racking, sorting room and market, and a
portrait layout serves phones. A URL switch composites the captured zones with
Spark \cite{spark}, using the four locally trained zones (8.0\,MB in total) in
place of the world-model captures of the same areas (5.9\,MB), and a separate
page compares each zone from its source cameras. The site runs from static
files, falls back to a local copy of the simulator when no backend is
configured, and was checked in desktop and phone layouts in headless Chromium.
Served locally to headless Chromium with software WebGL (SwiftShader, four CPU
cores, 1280\,$\times$\,720), the default view transferred 2.7\,MB and finished
loading in 1.5\,s, and the view with captures transferred 60\,MB in 2.7\,s.
Without a GPU it drew 1.1 and 0.5 frames per second, a floor rather than a
device figure; frame rates on laptops and phones with a graphics processor were
not measured.

\subsubsection{The OpenUSD stage renders outside the browser}
\begin{figure}[t]
\centering
\includegraphics[width=0.8\linewidth]{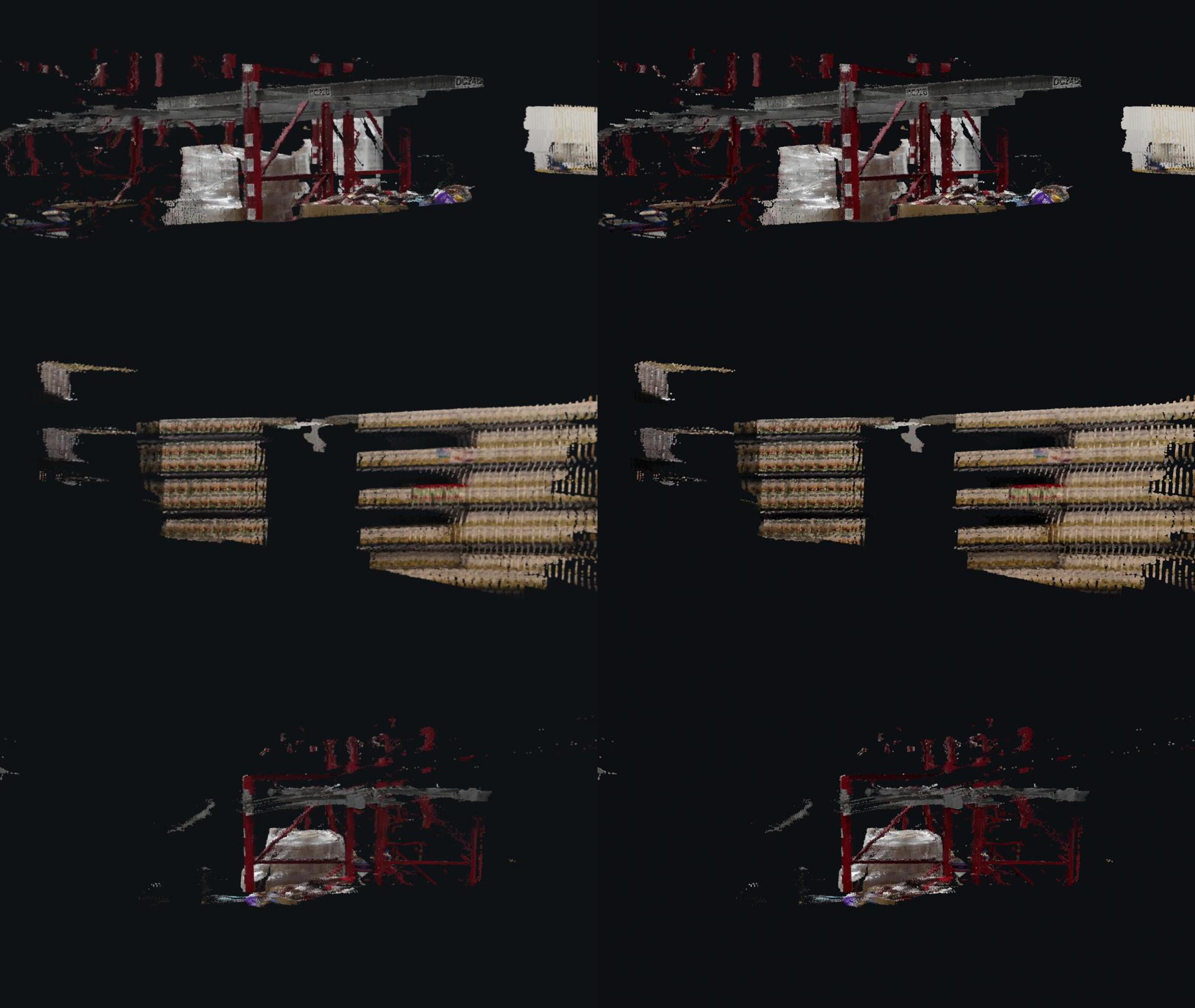}
\caption{The same OpenUSD stage rendered on a CPU, from identical cameras: the
Gaussian-splat renderer (left) and Blender 5.2 Cycles with one emissive sphere
per splat (right). Rows: the racking aisle, the client market, and the racking
aisle from an oblique viewpoint. All three shots are near the zones' source
viewpoints.}
\label{fig:usdrenders}
\end{figure}

\begin{figure}[t]
\centering
\includegraphics[width=\linewidth]{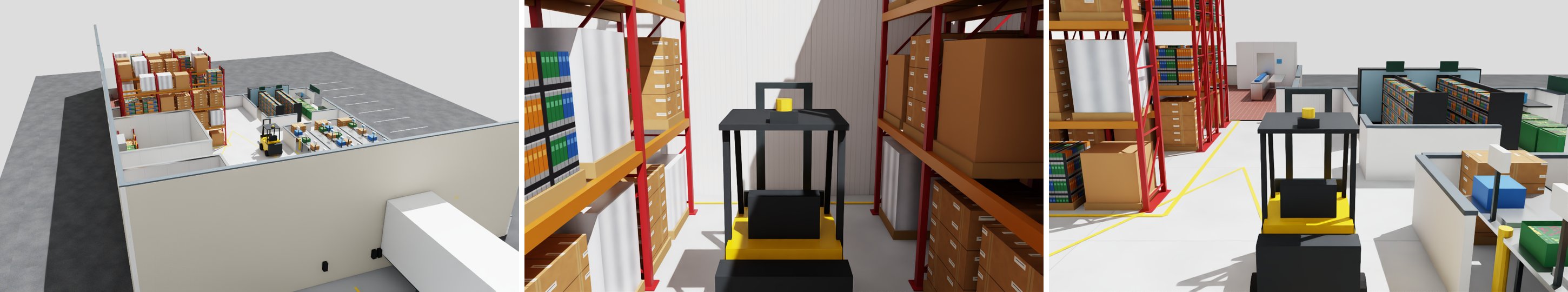}
\caption{The modeled facility rendered by NVIDIA Omniverse (Isaac Sim 5.1, Kit
107.3, RTX) on an EC2 g5.xlarge: an orbit over the building (left), the
forklift in a racking aisle (middle) and a chase view toward the receiving area
(right). Frames from the three clips of Movie~S3; the forklift follows a
scripted path, not live telemetry, and the splat captures are not shown
because this Kit version draws them blank.}
\label{fig:omniverse}
\end{figure}

Two renderers read \code{site.usdz} directly on a four-core CPU machine with
no GPU, from the same list of six camera shots (Figure~\ref{fig:usdrenders}). A
Gaussian-splat renderer written for this work walks the stage for every splat
prim, applies each prim's transform to positions and covariances, and
composites the splats in depth order; it rendered all six stills in about
20\,s including loading, and a 24\,s, 576-frame 1080p flythrough in a few
minutes. Blender 5.2's own USD importer brings in the transform hierarchy but
turns each splat prim into an empty object, because it does not read the
\code{ParticleField3DGaussianSplat} schema. Reading the splats with the OpenUSD
library that ships inside Blender and drawing each as an emissive sphere gave
renders in Cycles that match the splat renderer shot for shot, at about 5\,s
per 960\,$\times$\,540 frame with 8 samples. In both, each zone reads well only
near its source viewpoints; wide shots break into fragments because the
captures are sparse, which more frames or generated views would address, not
a different renderer, and they are why the modeled facility is the default
view. The modeled facility itself is ordinary meshes and materials written by
Blender's USD exporter, so any OpenUSD reader can open it.
Omniverse rendered the modeled facility on a rented GPU after six failed or
partial runs (Figure~\ref{fig:omniverse}, Movie~S3). On 25 October 2026 we ran
the Isaac Sim 5.1 container (Kit 107.3) headless on EC2 in us-east-2, each run
on an instance that terminated itself. On a g6.xlarge (one NVIDIA L4) with the
GPU image's 595-series driver, Kit crashed in its RTX scene database right
after start, even for a bare lit cube, so neither the twin files nor the splat
schema caused it. On plain Ubuntu with the 580-series driver the container is
validated on, Kit opened the stage. With no g6 capacity left in the region, the
final run used a g5.xlarge (one NVIDIA A10G) and rendered three 4\,s,
96-frame clips of the modeled facility at 1280\,$\times$\,720 with eight RTX
subframes per frame, in 18.6\,s, 47.7\,s and 49.4\,s, about 2 to 5 frames
per second offline. Kit 107.3 drew the splat captures' \code{ParticleField}
prims as blank grey, so Omniverse does not yet show the photoreal zones; the
browser and the CPU renderers do. The clips use a scripted forklift path, not
live telemetry. Instance boot took about 1 minute, pulling the container about
4.5 minutes, and Kit start-up 75\,s cold and 14\,s warm. Across seven runs the
trial used about 94 instance-minutes, about \$1.34 at on-demand list prices,
and left no instances or volumes behind. Interactive frame rate in Omniverse, and binding live telemetry to zone
prims there, were not measured; both need an interactive Kit session rather
than headless clip rendering.

\subsection{Cost and operational technology security}
What would it cost to run the twin on real floor data? We priced one
representative 135{,}000\,sq\,ft facility from list prices on 19 October 2026
(Table~\ref{tab:costest}); this is an estimate under stated assumptions, not
a measured bill. Battery LoRaWAN sensors report freezer, cooler, kitchen and
dock temperatures and door states through AWS IoT Core for LoRaWAN, BLE
beacons place each forklift in an area, the warehouse management system
supplies inventory, and the existing API serves the result with the
\code{simulated} flag cleared. About \$2{,}700 of hardware and about \$12 a
month of AWS services cover the core, so the cloud is not the barrier; the
hardware, its installation and keeping the master data current are. Running
Omniverse on an AWS GPU only when needed adds about \$124 a month, while an
always-on GPU would add about \$2{,}100, which is why the browser is the
everyday viewer. The Omniverse trial gives a measured reference point: seven
runs used about 94 instance-minutes on g6.xlarge (\$0.8048 per hour on demand)
and g5.xlarge (\$1.006 per hour) instances, about \$1.34 computed from list
prices because the bill lags by a day, and every instance was terminated with
no volumes left behind. Scaled up, on-demand rendering on ECS or AWS Batch
with a GPU minimum of zero costs about \$0.80 per GPU-hour actually used,
plus about 6 minutes (about 8 cents) of container pull and Kit start-up per
cold job; ten GPU-hours a week would be about \$35 a month on a g6.xlarge or
\$44 on a g5.xlarge, with no software fee for the container in this trial. Scaled to a complete capture of a 135{,}000\,sq\,ft facility with the same
tools (twelve zones, with world-model poses and generated views, one GPU-hour of
splat training per zone and ten hours of Omniverse review), the cloud spend from
video to deployed twin would be about \$25 to \$50 at list prices, before staff
time; the assumptions, unit prices and sources are in the repository's
\code{docs/COST\_ESTIMATE.md}.

The same principle, paying only while something runs, extends to every part of
the design and is drawn as a cost-controls panel in each architecture diagram
(Figs.~\ref{fig:aws}, \ref{fig:azure} and \ref{fig:gcp}). The API and ingest
functions already bill per request, so an idle twin costs almost nothing.
Work that is easier to ship as a container image, such as a warehouse import
that outlasts a function's time limit or a scene rebuild, can run on Amazon
Elastic Container Service (ECS) with AWS Fargate, which bills per second of
virtual CPU and memory with a one-minute minimum and costs nothing at zero
tasks; Azure Container Apps and Google Cloud Run do the same and scale to zero
on their own. Containers are not automatically cheaper: an always-on Fargate
task of 0.25 virtual CPU and 0.5\,GB on ARM costs about \$7 a month at
us-east-1 list prices, more than the whole demo API, so containers suit jobs
rather than the polling API. Fargate offers no GPUs, so Omniverse needs an EC2
GPU instance, started for a session and stopped afterward; Spot capacity
lowers its price further but can be reclaimed mid-session, and a stopped
instance's disk still bills. On AWS the same pattern can be automated by
running the Omniverse renderer as an ECS service or AWS Batch job on GPU EC2
capacity whose Auto Scaling group has a minimum of zero, so an instance exists
only while a render runs. A new AWS account starts with no on-demand GPU
quota, so a quota request is part of deployment. Azure Container Apps and
Cloud Run do offer serverless GPUs that scale to zero, which could host a
headless renderer; we have not tested Omniverse on them. A budget alert on
each cloud catches anything left running. All prices here are public list
prices, and the measurements come from an individual AWS account and the
quotas and policies it had. Enterprise accounts may differ through negotiated
discounts, savings plans, quotas and region policies, so the figures are
directional, not a quote.

\subsubsection*{Operational technology security}
Most security discussion around a cloud twin is about IT: the account, the
API and the web front end. A twin fed by floor sensors also brings operational
technology (OT) into scope: the sensors, the gateways that carry their data
and the path from the floor to the cloud. We therefore designed the twin's OT
boundary before any real device is connected (Figure~\ref{fig:aws}), following
the zone-based segmentation and Purdue levels described in NIST SP 800-82
Rev.~3 \cite{nist80082r3} and the zones-and-conduits model of the ISA/IEC 62443
series \cite{isa62443,iec62443_overview}.

The central decision is that the twin watches the floor and never drives it.
The API exposes only \code{GET} routes, and in the design no cloud principal may
publish to a device topic, so even a full compromise of the cloud account
could mislead people but could not change a freezer setpoint. A twin that
sends commands back would become part of the control system and would need
far more assurance. For the same reason the design adds its own battery
LoRaWAN sensors rather than tapping the refrigeration controllers, whose local
alarms stay authoritative. Floor devices and gateways (Purdue levels 0 to 2)
sit behind an OT DMZ (level 3.5) that allows only outbound TLS, and the only
designed path toward the floor is a signed, human-approved firmware update.

For a read-only twin the main OT risk is integrity rather than control. A
falsified ``all normal'' reading could hide a cold-chain failure, and
falsified alarms could train staff to ignore real ones. MITRE ATT\&CK for ICS
lists such falsified telemetry as an unauthorized reporting message
(T1692.002) \cite{mitre_t1692}, and NIST SP 800-82 notes that many field
devices and protocols allow data to be replayed, modified or spoofed
\cite{nist80082r3}. The design answers with one identity per device: an X.509
certificate for each MQTT device, with an AWS IoT policy that lets a device
connect only under its own registered thing name
\cite{aws_iot_connect_policy} and certificates that can be revoked permanently
\cite{aws_iot_revoke}, and per-device AES-128 session keys for LoRaWAN
sensors. Readings would then pass plausibility, rate-of-change and data-age
checks before they reach staff. One trust point should be stated plainly:
AWS IoT Core for LoRaWAN decrypts sensor payloads in the cloud
\cite{aws_lorawan_security}, so the cloud provider is inside the trust
boundary for that data.

What is built and what is designed must be kept apart. Implemented and
deployed today are the cloud-side controls: the \code{GET}-only HTTPS API, its
throttle of 5 requests per second (burst 10), a function role that can only
write its own logs, seven-day log retention, and no IoT resources or device
credentials in the account. Every device-side control above is a design,
because the demonstration uses simulated telemetry and no real device exists.
The target security levels are reasoned, not the result of an IEC 62443
assessment, and the public API has no authentication because the data it
serves is public and simulated; a deployment with real readings would add
both.

\begin{table}[t]
\caption{Estimated cost of running the representative twin on real floor
data, from public list prices checked on 19 and 26 October 2026. Excludes tax, installation
and staff time. Directional: enterprise discounts and quotas may differ.}
\centering\small
\begin{tabularx}{\linewidth}{@{}>{\raggedright\arraybackslash}Xrr@{}}
\toprule
Item & One-time & Monthly \\
\midrule
24 temperature sensors, 21 door sensors, 2 LoRaWAN gateways & \$1{,}850 & \\
8 forklift beacons, 10 BLE gateways (area-level location) & \$840 & \\
AWS: IoT Core, Lambda, API Gateway, DynamoDB, logs (20 viewers) & & \$12 \\
Optional: Omniverse on a g6e.xlarge, 10\,h a week & & \$124 \\
Measured: Omniverse trial, 7 runs, 94 instance-minutes\textsuperscript{a} & \$1.34 & \\
Scaled: on-demand rendering on g6.xlarge, 10\,h a week & & \$35 \\
Optional: forklift telematics, 8 vehicles & & \$200--400 \\
\bottomrule
\end{tabularx}
\par\footnotesize\textsuperscript{a}Computed from list price; the bill was not yet available.
\label{tab:costest}
\end{table}

\subsection{Methods}
\subsubsection{Cost controls for paid generation}
World-model calls are the only paid, non-idempotent steps, so they are
guarded in code. On the Marble 2 route, each paid stage records its operation
identifier before waiting and sends an idempotency key derived from the request
body, so a re-run resumes or reuses a result instead of paying again; a credit
ceiling is passed per stage. The Marble v1 route adds the following guards.
Only the draft model is allowed unless explicitly overridden, and the model is
checked again in the payload and in the returned world. Each process makes at
most one billable call. The operation identifier is saved the moment a job is
accepted, so a crash resumes polling instead of paying again, and a finished
world is reused unless a new one is forced. A lifetime ledger refuses any run
that would exceed the budget, a live balance check requires a reserve, and the
billable request is retried only on HTTP 429, which the API defines as not
accepted, never on timeouts or server errors. These guards were exercised
against a local mock of the API covering a 429 retry, a 503 during polling,
resume in flight with exactly one generation across three runs, reuse,
refusal of non-draft models and ledger accounting.

\subsubsection{Serverless data backend}
The backend is one AWS Serverless Application Model template \cite{awssam}: an
S3 bucket serving the twin assets and web viewer; a Python 3.12 Lambda
function on arm64 with 128\,MB of memory and a 3\,s timeout behind an HTTP API
with \code{GET /telemetry}, \code{GET /operations} and \code{GET /health}, throttled to 5 requests per
second (burst 10) as a cost ceiling; and a log group with seven-day retention.
The telemetry function is stateless and deterministic: every value is a
function of the wall clock, so clients in the same second see the same state
without shared storage. Zone temperatures follow a slow sinusoidal drift with
a 15-minute period plus per-5-second jitter, doors open in about 10\% of
30-second windows, and alarm status is set when the deviation from setpoint
exceeds the zone's band. Responses are cacheable for 2\,s and every payload
carries \code{simulated: true}. Zone and asset anchors are in meters in the
twin frame, the frame the modeled facility is built in, so labels sit on the
modeled cold rooms by construction. Each capture's placement and crop box are
set by hand in a configuration file; their residual error against the real
building is not measured, because no survey exists.

\subsubsection{Operations model}
The master data lists areas with footprints and positions in meters, six
inventory families with a home temperature zone, pallet weight, shelf life and
days-of-supply target, five inbound source types and four outbound channels
with their shares and product mixes, and the kitchen's programs, equipment,
recipes with per-meal bills of materials, and routing. Each value is either
tied to a reference or marked as an estimate. Daily inbound follows the annual
volume, a weekday profile and deterministic noise; outbound is set so stock
tracks a slow target curve, so one day's closing stock is the next day's
opening stock. Areas whose cold-chain sensor is in alarm report their stock as
at risk; the kitchen walk-in is a fourth cold-chain sensor in the telemetry
feed. Kitchen batches of 250 meals start at 06:00 local time, staggered by 45
minutes, and are dispatched between 13:00 and 15:30. Tests check that the
Python and JavaScript payloads agree, that stock carries over midnight and that
Sunday is closed. The weekday ranges in Results were computed by evaluating the
model at the end of each day from 1 September to 26 October 2026.

\subsubsection{Visualization}
The modeled facility is built by \code{scripts/build\_facility.py} with the
\code{bpy} module of Blender on Python 3.13, without a display. Footprints
come from the operations master data; materials use small procedural textures
generated by the script, so the build needs no downloads. One mesh per area and
material keeps the browser to about 150 draw calls. The script exports the
scene as binary glTF (Y up, transforms applied) and as USDZ with materials, and
can also save a \code{.blend} file and a Cycles still (48 samples, denoised,
CPU). A test checks that the glTF file carries every operational area and the
forklift node and that the building floor spans the telemetry frame.

The stage to open in Omniverse is \code{twin.usda}. A script writes it from
the browser configuration: a reference to the modeled facility, one reference
per capture with the viewer's translation and rotation, a crop sublayer that
sets the opacity of every splat outside the capture's crop box to zero, and a
reference to \code{operations.usda}; a test compares the stage's placements
with the configuration. A second script renders the captures on the CPU with
the area footprints drawn over them as a visual check. In Omniverse, a small
extension is planned that polls \code{/telemetry} and updates prims under the
twin: zone markers change color with alarm status, labels show temperatures,
and the forklift follows its reported position and heading. The clips in
Movie~S3 were rendered by \code{scripts/omniverse\_render\_clips.py} inside the
Isaac Sim 5.1 container, launched by \code{scripts/gpu\_trial.py} on an
instance that uploads its output through presigned S3 URLs and then
terminates itself. Because \code{/Facility} carries the Z-up to Y-up rotation,
the forklift's world pose is mapped into that frame before it is set.

The browser viewer uses three.js 0.180 and its glTF loader for the modeled
facility, and Spark 2.2 \cite{spark} only when the captures are switched on.
It deploys as static files to GitHub Pages; common USD web viewers hand USDZ
to AR Quick Look, which does not render USD particle fields. The CPU splat
renderer uses usd-core, NumPy and Numba, projects each Gaussian with an
elliptical weighted average filter, colors it from its degree-0 spherical
harmonics and blends front to back. For the captures in Blender, a script
reads every splat prim with the \code{pxr} module, builds one point cloud per
prim with Geometry Nodes (radius 1.7 times the splat scale, emission color,
opacity as transparency) and renders with Cycles on the CPU. Spheres are not
Gaussians, so edges are harder than in the splat renderer.

The architecture diagrams (Figure~\ref{fig:aws} and
Appendix~\ref{app:clouds}) are generated as draw.io files by
\code{paper/diagrams/make\_diagrams.py} with the AWS, Azure and Google Cloud
icon libraries of draw.io 26.2.15 and exported to PDF with draw.io desktop.

\section{Equivalent designs on Microsoft Azure and Google Cloud}
\label{app:clouds}
\begin{sidewaysfigure}
\centering
\includegraphics[width=\linewidth]{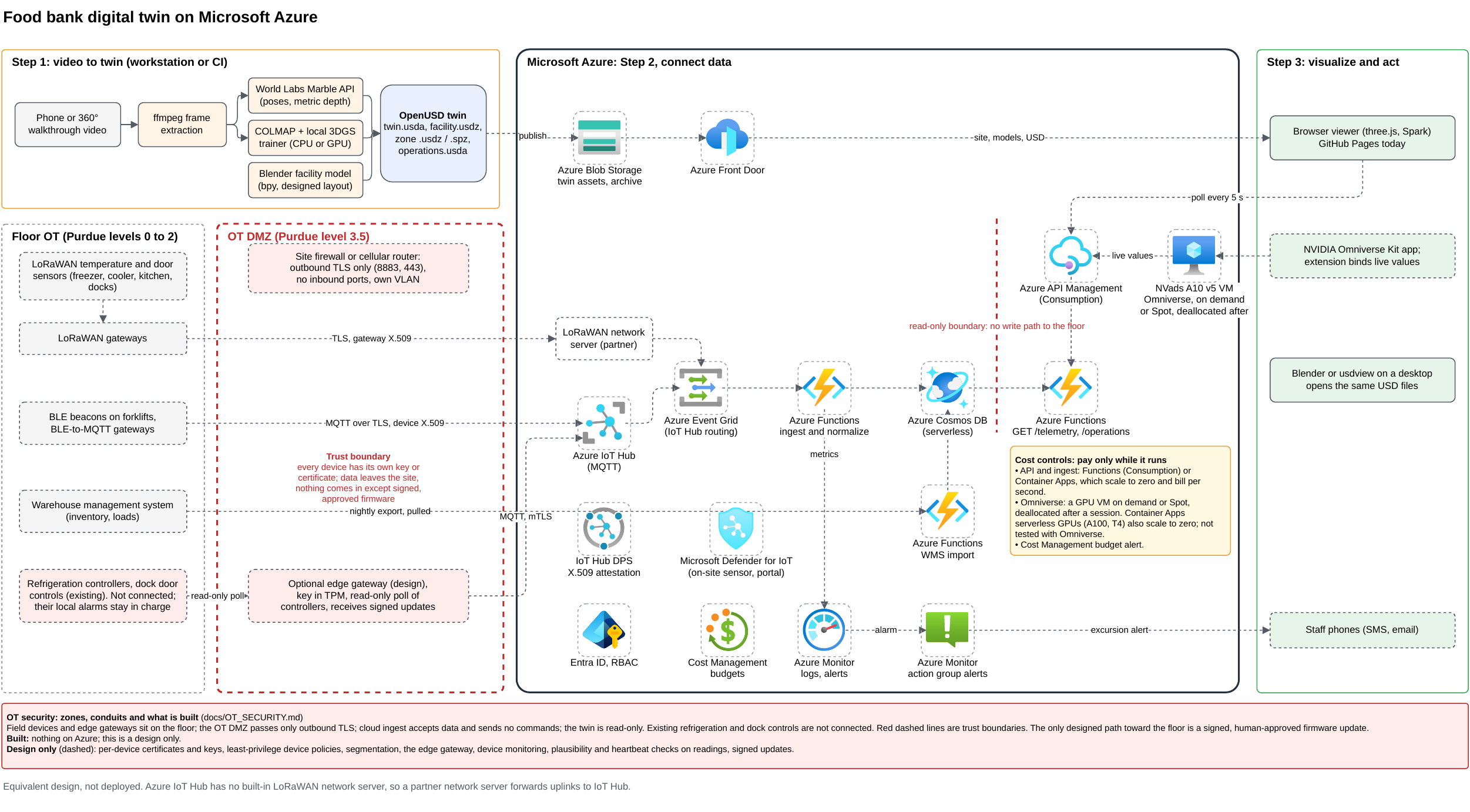}
\caption{The food bank twin on Microsoft Azure: an equivalent design, not
deployed, with the same OT zones and trust boundaries as Figure~\ref{fig:aws}. Drawn in draw.io with the Azure icons.}
\label{fig:azure}
\end{sidewaysfigure}

\begin{sidewaysfigure}
\centering
\includegraphics[width=\linewidth]{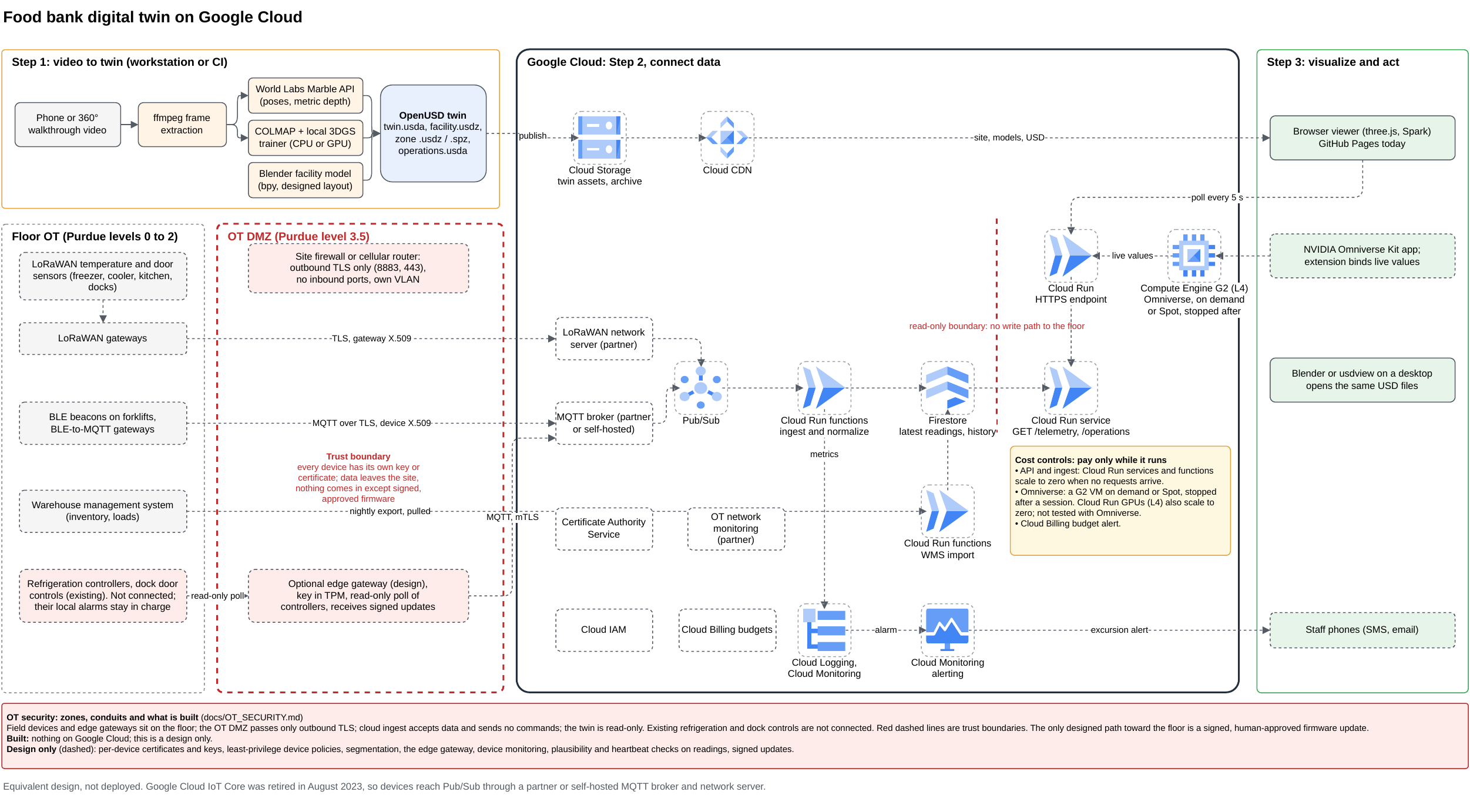}
\caption{The food bank twin on Google Cloud: an equivalent design, not
deployed, with the same OT zones and trust boundaries as Figure~\ref{fig:aws}. Drawn in draw.io with the Google Cloud icons.}
\label{fig:gcp}
\end{sidewaysfigure}

Nothing in the pipeline depends on AWS: the OpenUSD twin is files, and the API
is a stateless function behind an HTTPS endpoint. Figures~\ref{fig:azure}
and~\ref{fig:gcp} redraw Figure~\ref{fig:aws} with the same layout on the
other two large clouds, service for service (Table~\ref{tab:clouds}). Neither
has been deployed or priced. Two differences matter for a food bank. Azure IoT
Hub has no built-in LoRaWAN network server, so a partner network server
forwards the sensors' uplinks to it. Google Cloud IoT Core was retired in
August 2023, so on Google Cloud the sensors and gateways reach Pub/Sub through
a partner or self-hosted LoRaWAN network server and MQTT broker.

\begin{table}[h]
\caption{The same twin on three clouds.}
\centering\small
\begin{tabularx}{\linewidth}{@{}>{\raggedright\arraybackslash}p{0.2\linewidth}>{\raggedright\arraybackslash}X>{\raggedright\arraybackslash}X>{\raggedright\arraybackslash}X@{}}
\toprule
Role & AWS & Azure & Google Cloud \\
\midrule
LoRaWAN network server & IoT Core for LoRaWAN & Partner network server & Partner network server \\
MQTT ingest & IoT Core & IoT Hub & Partner or self-hosted broker \\
Routing & IoT Core rules & Event Grid & Pub/Sub \\
Functions & Lambda & Functions & Cloud Run functions \\
Readings store & DynamoDB & Cosmos DB (serverless) & Firestore \\
Object storage & S3 & Blob Storage & Cloud Storage \\
HTTP API & API Gateway HTTP API & API Management (Consumption) & Cloud Run \\
Content delivery & CloudFront & Front Door & Cloud CDN \\
Monitoring, alerts & CloudWatch, SNS & Azure Monitor, action groups & Cloud Logging, Cloud Monitoring \\
Access, budgets & IAM, Budgets & Entra ID and RBAC, Cost Management & Cloud IAM, Billing budgets \\
GPU for Omniverse & EC2 g6e.xlarge (L40S) & NVads A10 v5 & Compute Engine G2 (L4) \\
Containers, pay per second & ECS on Fargate (no GPU) & Container Apps (scale to zero) & Cloud Run (scale to zero) \\
Serverless GPU & none; EC2 on demand or Spot & Container Apps serverless GPUs & Cloud Run GPUs (L4) \\
\bottomrule
\end{tabularx}
\label{tab:clouds}
\end{table}

\clearpage
\printbibliography

\end{document}